%% file: main.tex
\documentclass[sigconf]{acmart}

\usepackage{graphicx}
\usepackage{float}
\usepackage{amsfonts} 
\usepackage{amsmath}
\usepackage{subfig}
\usepackage{bm}
\usepackage{mathrsfs}
\usepackage{booktabs}
\usepackage{multirow}
\usepackage{caption}
\usepackage{enumitem}

\newcommand{\bigCI}{\mathrel{\text{\scalebox{1.07}{$\perp\mkern-10mu\perp$}}}}

\input{macros}

\setcopyright{none}

\renewcommand\footnotetextcopyrightpermission[1]{} 
\begin{document}
\title{Assessing the Causal Impact of COVID-19 Related Policies on Outbreak Dynamics: A Case Study in the US}

\author{Jing Ma$^1$, Yushun Dong$^1$,  Zheng Huang$^1$, Daniel Mietchen$^{1,2}$, Jundong Li$^{1,2}$}

\affiliation{
  {$^1$University of Virginia, Charlottesville, VA \country{USA} 22904}\\ 
  {$^2$Global Infectious Disease Institute, University of Virginia, Charlottesville, VA \country{USA} 22904}\\
    \{jm3mr, yd6eb, zh4zn, dm7gn, jundong\}@virginia.edu
}


\begin{abstract}

To mitigate the spread of COVID-19 pandemic, decision-makers and public authorities have announced various non-pharmaceutical policies. Analyzing the causal impact of these policies in reducing the spread of COVID-19 is important for future policy-making. The main challenge here is the existence of unobserved confounders (e.g., vigilance of residents).
Besides, as the confounders may be time-varying during COVID-19 (e.g., vigilance of residents changes in the course of the pandemic), it is even more difficult to capture them.
In this paper, we study the problem of assessing the causal effects of different COVID-19 related policies on the outbreak dynamics in different counties at any given time period. 
To this end, we integrate data about different COVID-19 related policies (treatment) and outbreak dynamics (outcome) for different United States counties over time and analyze them with respect to variables that can infer the confounders, including the covariates of different counties, their relational information and historical information. 
Based on these data, we develop a neural network based causal effect estimation framework which leverages above information in observational data and learns the representations of time-varying (unobserved) confounders.
In this way, it enables us to quantify the causal impact of policies at different granularities, ranging from  a category of policies with a certain goal to a specific policy type in this category. 
Besides, experimental results also indicate the effectiveness of our proposed framework in capturing the confounders for quantifying the causal impact of different policies. More specifically, compared with several baseline methods, our framework
captures the outbreak dynamics more accurately, and our assessment of policies is more consistent with existing epidemiological studies of COVID-19. All the data and code can be accessed via \href{https://github.com/QIDSOD/COVID-19-Policy-Causal}{https://github.com/QIDSOD/COVID-19-Policy-Causal}.

%

\end{abstract}

%


 \maketitle

\input{intro}

\input{data}
\input{method}
\input{exp}

\input{related}

\vspace{-2mm}
\section{Conclusion} 
In this paper, we study the problem of assessing the causal impact of various COVID-19 related policies on the outbreak dynamics in different U.S. counties at different time periods throughout 2020. The main challenge 
here
is the existence of unobserved and time-varying confounders. 
To address this problem, we 
integrate
data from multiple COVID-19 related data sources 
containing
different kinds of information which can serve as proxies for confounders. We develop a neural network based framework which learns the representations of the confounders by utilizing 
relational and time-varying observational data 
and then estimates the causal effect of these polices on the outbreak dynamics with the learned confounder representations. Based on the estimated causal effects, it enables the assessment of these policies at different granularities. We also compare the prediction performance of outbreak dynamics and the presence of policies, as well as the causal effect estimation results of our framework with other baselines. The results implicitly validate the capability of our framework in controlling confounders for causal assessment of COVID-19 related policies.



\clearpage
\bibliographystyle{ACM-Reference-Format}
\bibliography{ref}

\end{document}

%% file: macros.tex
\newtheorem*{conjecture*}{Conjecture}
\newtheoremstyle{nonindented}{1ex}{1ex}{}{}{\bfseries}{.}{.5em}{}
\newtheoremstyle{indented}{1ex}{1ex}{\itshape\addtolength{\leftskip}{0.6cm}\addtolength{\rightskip}{0.6cm}}{}{\bfseries}{.}{.5em}{}
\theoremstyle{nonindented}
\theoremstyle{indented}
\theoremstyle{plain}

\renewcommand{\hat}{\widehat}
\renewcommand{\tilde}{\widetilde}

\def\max{\qopname\relax n{max}}

\newcommand{\eat}[1]{}

\newenvironment{lp*}{\begin{equation*}  \begin{array}{lll}}{\end{array}\end{equation*}}


%% file: intro.tex
\section{Introduction}

The coronavirus disease 2019 (COVID-19) has spread quickly across the world, seriously affecting human health \cite{goodman2020using,li2020impact}, economics \cite{mckibbin2020economic,hevia2020conceptual}, and even politics \cite{romano2020covid19,misra2020psychological}. 
To address and mitigate the spread of the disease, political decision-makers and public authorities have issued various policies aimed at limiting the impact of the COVID-19 pandemic on different aspects of societal life within their remit \cite{hale2021global,hsiang2020effect,balmford2020cross,kayi2020policy,alon2020should}. 
Nevertheless, different policies may contribute differently to combating COVID-19 \cite{kucharski2020effectiveness,goolsbee2021fear}. 
Correspondingly, a natural question to ask is: \emph{which policy is more effective to mitigate the spread of COVID-19 in a given context}? 
There have been various studies that address this question from a statistical perspective, such as those using correlation analysis \cite{miller2020correlation,jia2020modeling}.
Such studies can capture statistical dependencies between the policies and the spread of COVID-19, but not identify the causal relationships between them. 
Yet answering this question from a \textit{causal} perspective is essential, as it can provide guidance to policymakers for addressing other pandemics or even further waves of the current one. However, the gold-standard of causal effect estimation, i.e., a randomized controlled trial comparing the outcomes of different treatments \cite{chalmers1981method}, is not readily applicable to study the causal impact of policies on the outbreak dynamics under pandemic circumstances, due to a range of ethical, legal and practical issues \cite{adebamowo2014randomised}. 
Hence, the assessment of the causal impact of different policies on the COVID-19 outbreak dynamics (e.g., the numbers of confirmed cases) is expected to be directly conducted from the observational data. Fortunately, it is easy to amass a large amount of observational data (e.g., whether a specific policy is in effect in a given county, the number of confirmed cases in that county) over time to support such studies. 

%

In this paper, we focus on assessing the causal impact of different COVID-19 policies on the outbreak dynamics with observational data. More specifically, we study this problem: given a specific time period, what is the causal effect of COVID-19 policies (\emph{causes/treatments}) on the outbreak dynamics (\emph{outcomes}) in each county (\emph{instance})? For example, in March 2020, how would the number of confirmed cases have been different in Albemarle county in Virginia \textit{because of} the state government having enacted social distancing policies?
One key challenge of conducting such causal effect estimation from observational data is the existence of \textit{unobserved confounders} (\textit{confounders} are the factors influencing both the \textit{treatment}, e.g., whether a policy is in effect, and the \textit{outcome}, e.g., the number of confirmed cases), which may bring \emph{confounding bias} \cite{rubin2005bayesian} into the estimation process. 
For example, in a county where residents have a high vigilance towards COVID-19, the government may issue policies to enforce 
social distancing
at an early stage of the pandemic, but residents in this county also tend to be more alert to COVID-19 and thus will have lower probability of infections, even in the absence of said policies. In this example, vigilance of residents is a confounder, and if it is not handled properly, we may incorrectly take the statistical correlations between the presence of these policies and the outbreak dynamics (e.g., in terms of number of confirmed cases) as a causal relation. 

Confounding bias can be eliminated by adjusting for all the confounders (i.e., controlling for confounders) \cite{rubin2005bayesian}. However, most of the existing works \cite{hevia2020conceptual,mcbryde2020role,ozili2020covid} which assess the causal effects of COVID-19 related policies on the outbreak dynamics from the observational data are mainly based on the unconfoundedness assumption \cite{rubin2005bayesian} (i.e., all the confounders can be observed). 
Similarly, the widely used difference-in-differences (DID) methods \cite{abadie2005semiparametric} are based on the parallel trend assumption \cite{goodman2020using}, which assumes that there are no (unobserved) factors influencing both the treatment and the growth trend of the outcome, i.e., the change of the outcome over a time period. 
However, in real-world scenarios, these assumptions are difficult to be satisfied because there often exist unobserved confounders. In the previous example, residents' vigilance could be a confounder, and it is hard to be quantified. 
Without effective ways to control for confounders, the aforementioned methods may yield biased estimation results \cite{rubin2005bayesian,CFR,causalForest}. 
Furthermore, since the COVID-19 pandemic has lasted over a year, the COVID-19 related observational data is naturally dynamic and evolving over time (e.g., when different policies are in effect in each county, and the number of confirmed cases in each county over time). One new challenge here is that the unobserved confounders may also be time-varying, e.g., residents' vigilance may be low at an early stage, but increases when the situation becomes more severe. Taking advantage of the temporal information could help us better capture and adjust for the time-varying unobserved confounders. Nevertheless, to the best of our knowledge, most of the current studies on COVID-19 related policies lack such capability~\cite{gencoglu2020causal,CFR}. 

To remedy the above introduced issues, 
we first adopt a weaker form of the unconfoundedness assumption~\cite{abadie2005semiparametric}, which enables us to capture the unobserved confounders from the proxy variables for them, i.e., the variables which have dependencies with the unobserved confounders. For example, certain confounders such as residents' vigilance in a county can be inferred from the popularity of searches about COVID-19 related keywords on 
Google Trends. Besides, residents' vigilance can also be inferred from the relational information among different counties, such as a county-to-county distance network. One potential reason is that neighboring counties tend to have more interactions and similar culture, thus their residents will have similar levels of vigilance. Historical information such as the adopted policies and the spread of COVID-19 at earlier time periods may also influence the current confounders such as residents' vigilance.
With the aid of such proxy variables, 
the confounders can be captured and thus an unbiased causal effect estimation becomes possible. Bearing this in mind, we 
integrate data
from several different data sources, 
covering 391 counties in the United States. More specifically, 
our data
includes COVID-19 related policies as treatments, the number of confirmed cases or death cases as outcomes. Besides, it also includes multiple covariates for each county and relational information among different counties, which can serve as proxy variables for capturing unobserved confounders. The data we collected spans from January to December 2020. 
To tackle our studied problem, we utilize the above observational data and propose a neural network based framework to learn the representations of (unobserved) confounders. Based on the learned representations, we then conduct causal effect estimation of different policies on COVID-19 outbreak dynamics. More specifically, in our proposed framework, a Recurrent Neural Network (RNN) \cite{medsker2001recurrent} is utilized to capture temporal information, while a Graph Convolutional Network (GCN) \cite{kipf2016gcn} is used to handle the relational information among different counties. At each time period, we predict the outbreak dynamics and whether these policies are in effect in each county. In this way, we can effectively take advantage of the observational data and learn the representations of time-varying confounders for causal effect estimation of COVID-19 policies on the outbreak dynamics.
Our contributions can be mainly summarized as:
\begin{itemize}[topsep=5pt]
\setlength\itemsep{-0.0em}
    \item We study 
    the important problem of estimating the causal effect of COVID-19 policies on the outbreak dynamics for a given county and time period. 
    \item We integrate data from several COVID-19 related sources. 
    \item We develop a neural network based framework based on both relational and time-varying observational data to control the influences of unobserved confounders.
    \item We assess the causal effect of different policies on the outbreak dynamics of COVID-19 with the developed framework. Specifically, we conduct the assessment for policies at different granularities, ranging from a category of policies with a certain goal to a specific policy type in this category. 
    \item We conduct 
    experiments to evaluate our framework. We find that our framework outperforms several alternatives and also 
    yields
    important insights for current or future pandemics.
\end{itemize}

%% file: data.tex
\vspace{-2mm}
\section{Data and Analysis}
In this section, we describe how we prepare 
data for
assessing the causal impact of COVID-19 related policies on 
outbreak dynamics. Some preliminary data analyses are also presented to show the potential capability of 
capturing the (unobserved) confounders.

\begin{figure}[!t]
	\centering  
	\includegraphics[width=\linewidth]{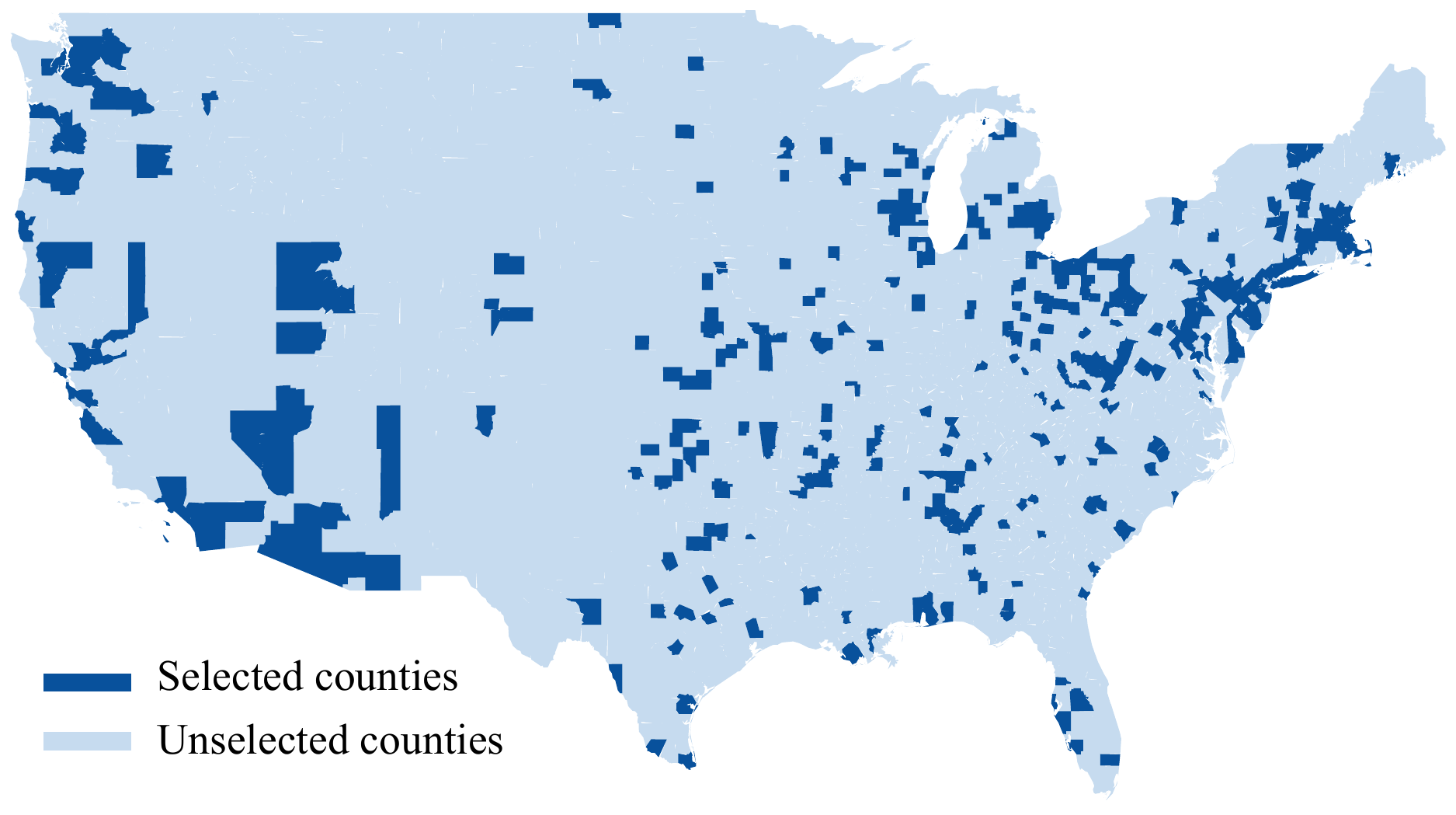}
	\caption{Geolocation of the selected counties in our 
	corpus.} 
	\label{map}  
	\vspace{-0.5cm}
\end{figure}

\subsection{Observational Data}
In general, two basic types of information are indispensable in the causal inference study, i.e., treatment and outcome. Specifically, for treatment, we collect COVID-19 related policies that 
have been enacted
by different counties in the United States throughout 2020; for outcome, we use the numbers of confirmed cases and death cases of different counties throughout 2020. To control the influence of unobserved confounders, we also collect data regarding the covariates of different counties and their relations. In particular, two types of networks among counties are used in our study. In total, after filtering the counties without sufficient data, 
our data corpus
includes 391 counties in the United States. The locations of these selected counties in our corpus are shown in Fig. \ref{map}. We then introduce how we collect and preprocess the data as follows.

\begin{figure*}[!t]
	\centering 
    \subfloat[Social distancing]{
	\label{effect_confirm_RO}
        \includegraphics[width=0.33\textwidth,height=1.7in]{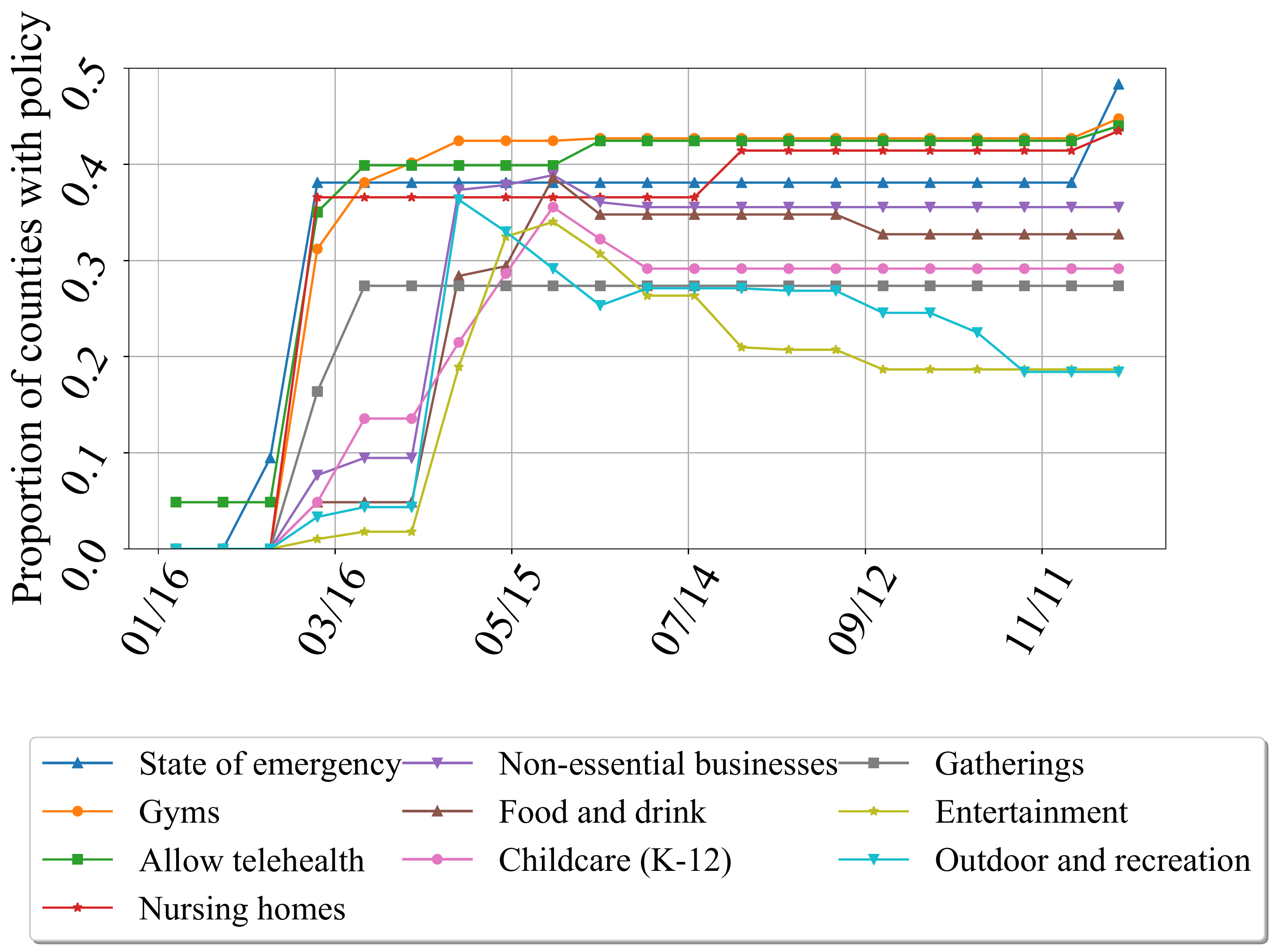}
    } 
    \subfloat[Reopening]{
    	\label{effect_death_RO}
        \includegraphics[width=0.33\textwidth,height=1.7in]{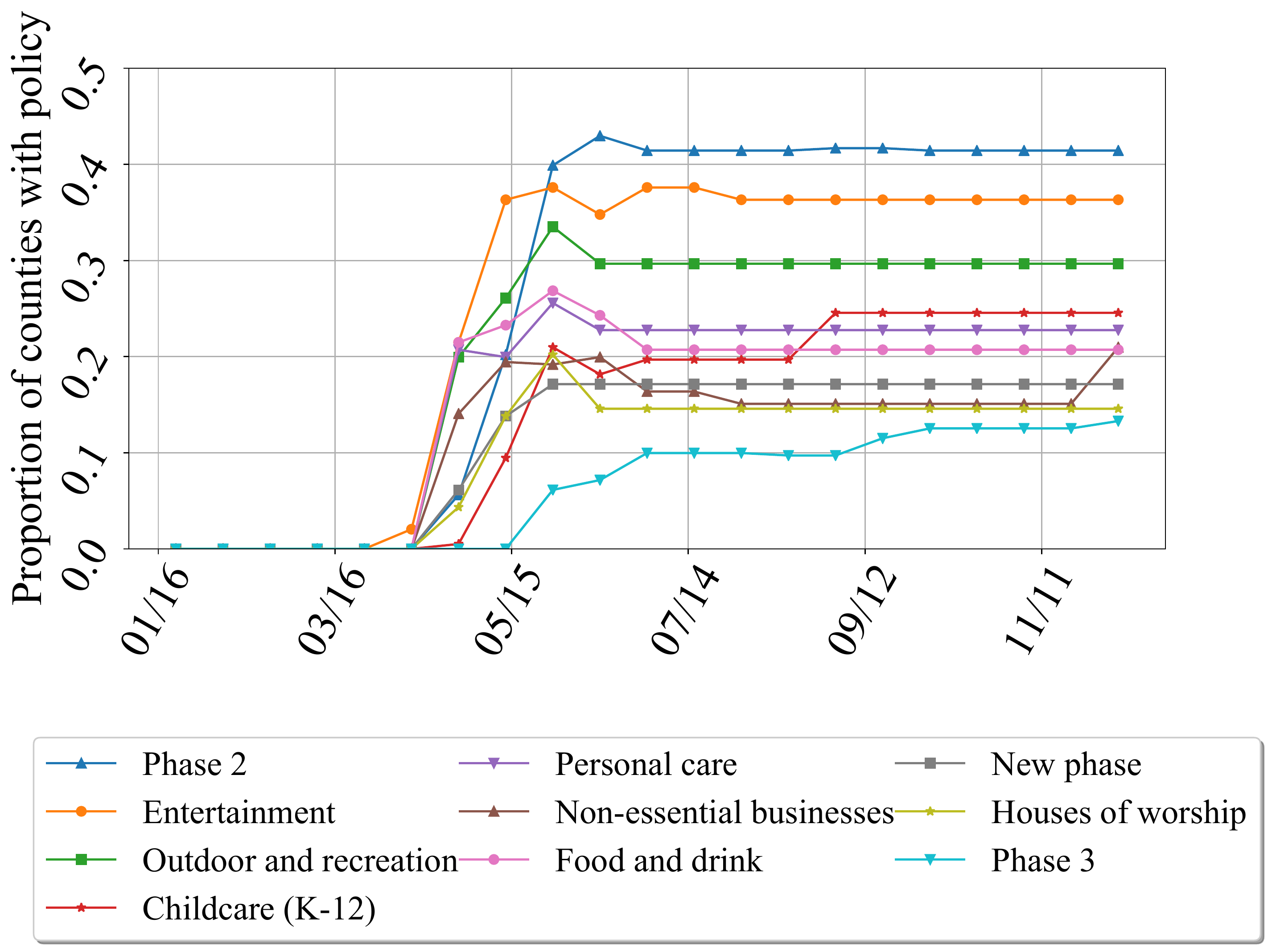}
    }
    \subfloat[Mask requirement]{
    	\label{assign_RO}
        \includegraphics[width=0.32\textwidth,height=1.7in]{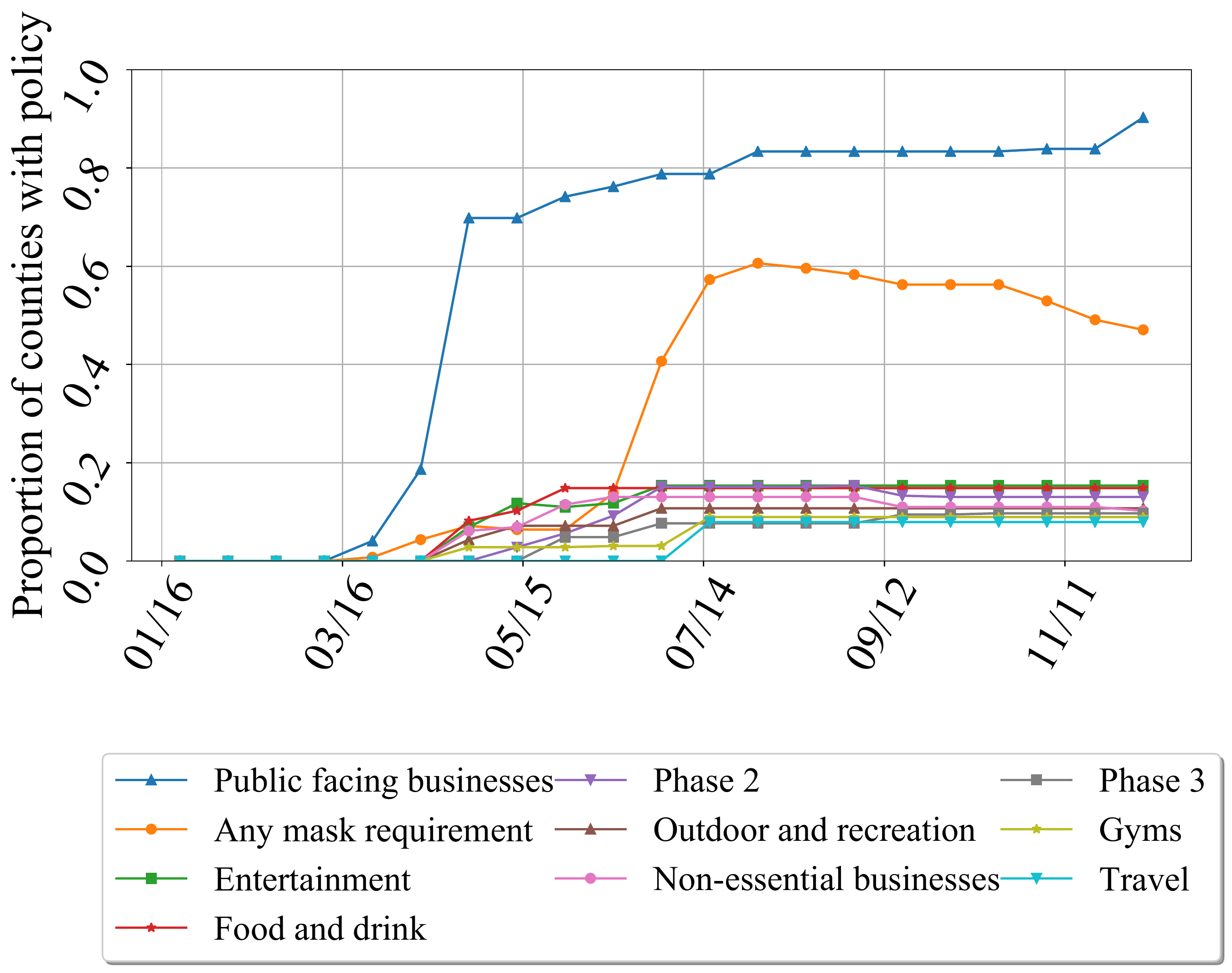}
    }
	\caption{Proportion of counties with policy types in each category over the course of 2020.}
	\label{fig:assignment}
	\vspace{-0.4cm}
\end{figure*}

\noindent\textbf{Treatment --- COVID-19 related policies.}
We collect the COVID-19 related policies that are in force in the USA throughout 2020 from the Department of Health \& Human Services\footnote{https://catalog.data.gov/dataset/covid-19-state-and-county-policy-orders}. A total number of $60$ 
policy types
are included, along with descriptions and start/end dates.
The collected policies include both state-level and county-level ones. For state-level policies, we assume that 
the policy applies to all counties in the state;
for county-level policies, they are considered as only 
applying to
the corresponding county.
In order to better analyze the effect of these policies, we perform the following preprocessing:
(1) \textit{Policy filtering.} We remove the policy types that are adopted in less than $10\%$ of 
the counties in our corpus,
as we do not have sufficient observational data 
with respect to them.
%
%
(2) \textit{Policy categorization.} Based on the goal and some key element of each policy, we group them into three categories: \emph{reduce contacts through social distancing} (henceforth referred to as social distancing), \emph{minimize damage to the economy through reopening} (reopening), and \emph{reduce airborne transmission through mask requirements} (mask requirements). 
For each category, Fig.~\ref{fig:assignment} shows the top-$10$ policy types adopted by the largest number of counties, and 
the proportion of counties adopting them throughout 2020.

\noindent\textbf{Outcome --- numbers of confirmed and death cases.}
The daily numbers of confirmed and death cases are collected across these $391$ counties from Johns Hopkins Coronavirus Resource Center\footnote{https://coronavirus.jhu.edu/map.html} 
from January to December, 2020. In our experiments, we regard 
these numbers
as outcomes of each county.

\noindent\textbf{Covariates --- popularity of search keywords on Google.} 
Unobserved confounders which causally affect the policies and outbreak dynamics are hard to 
capture.
Hence, we use some proxy variables such as covariates of counties to infer these confounders. More specifically, we consider the search of COVID-19 related keywords (e.g., coronavirus, mask, quarantine, etc.) by residents in these counties as 
covariates.
We collect the corresponding data from Google Trends\footnote{https://trends.google.com/trends/?geo=US}. 
In this process, we first select a set of COVID-19 related keywords, and then compute their \textit{popularity score} based on the corresponding proportion in the total searches in each county. A higher popularity score implies that a larger proportion in this county 
has a 
high
vigilance of COVID-19. In total, 19 different keywords are selected, and we obtain a 19-dimensional covariate vector for each county on each day 
from February to December 
2020.

\noindent\textbf{Networks --- distance network and mobility flow network.} 
Previous works \cite{netDeconf,guo2020ignite} have shown that network structure among instances can reflect some unobserved confounders. Therefore, in this work, two networks including the geographical distance network and mobility flow network among the selected 391 counties are collected as another kind of proxies for confounders.
\textit{1) Geographical distance network.}
Geographical distance network among counties can be utilized to capture confounders. Intuitively, counties that are geographically closer are more likely to have similar confounders~\cite{netDeconf} such as residents' vigilance, because they tend to have similar cultural background and social climate.
We construct a weighted network among the 391 counties based on the geographical distance from County Distance Database\footnote{https://www.nber.org/research/data/county-distance-database}, where nodes represent counties, and edge weights are calculated from the corresponding distances between county pairs. Specifically, we select the county pairs with distance less than a threshold $\tau=100$ kilometers, and set the  weight as $1/d(i,j)$ for the edge between the $i$-th and $j$-th counties with distance $d(i,j)$.
\textit{2) Mobility flow network.}
The mobility flow network among counties can also be adopted to capture confounders, as counties with large mobility flow are more likely to have similar confounders~\cite{netDeconf,guo2020ignite} (such as residents' vigilance) as they have more communications.
We construct an temporal mobility flow network with weighted edges among the 391 counties based on COVID19 USFlows~\cite{kang2020multiscale}, which tracks anonymous GPS pings based on mobile applications. 
In this temporal network, the total daily volume of mobility flow is aggregated at different scales (e.g., census tract, county, and state) w.r.t. the timeline spanning from February to December in 2020. Each node denotes a county, and the weight of each edge is calculated from the mobility flow volume between the corresponding pair of counties. Specifically, we set the weight as $\frac{\log{flow(i,j)}}{\max_{i,j}\log{flow(i,j)}}$ for the edge between the $i$-th and $j$-th counties with mobility flow $flow(i,j)$.



\begin{figure*}[!htbp]
	\centering 
	\subfloat[Heatmap: Pearson correlation of \\confirmed case number between counties.]
	{
	\label{pear-a}
        \includegraphics[width=0.264\textwidth]{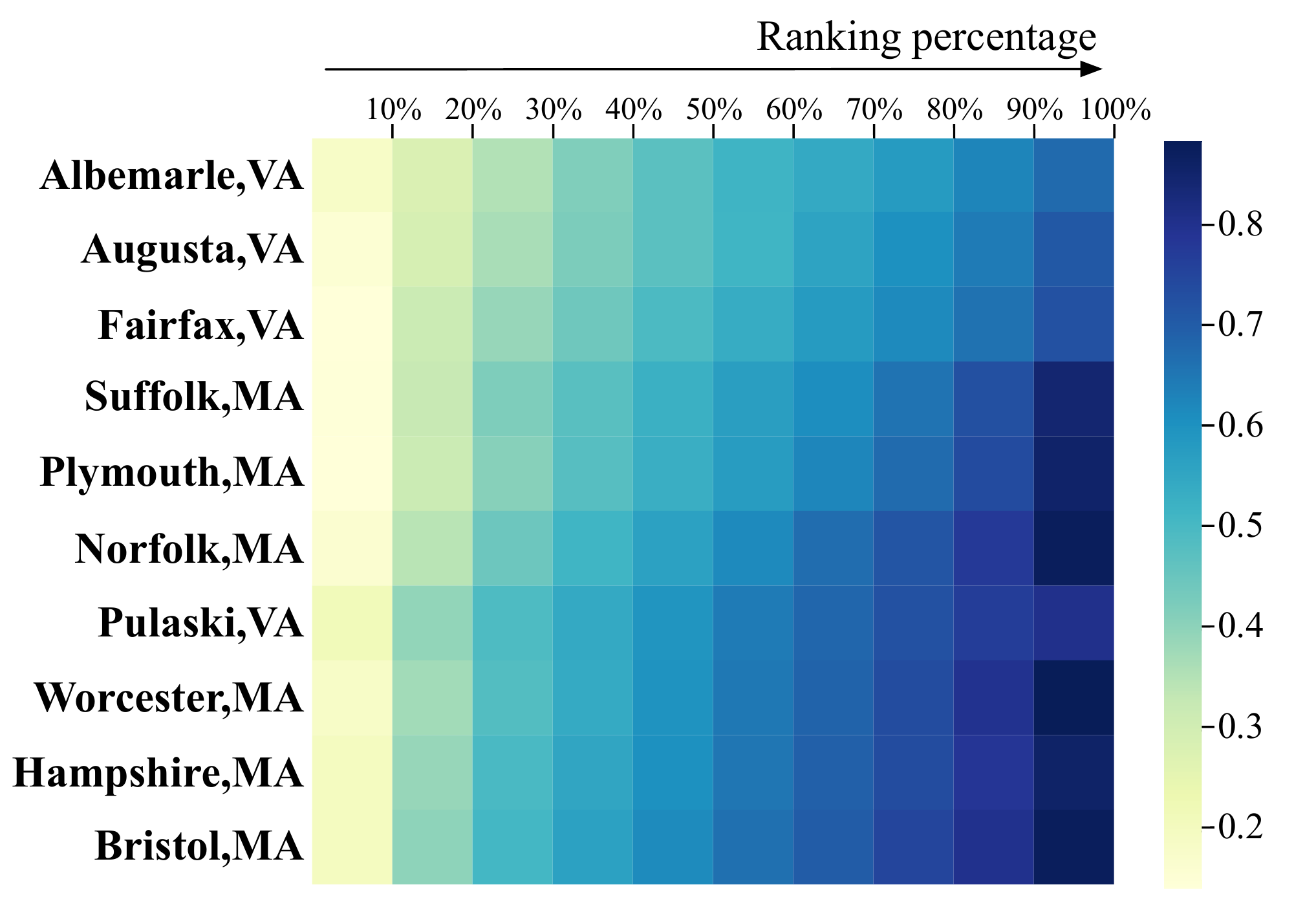}
    } 
    \hspace{3mm}
    \subfloat[Heatmap: Pearson correl-\\ation of keyword popularity.]
    {
    	\label{pear-d}
        \includegraphics[width=0.207\textwidth]{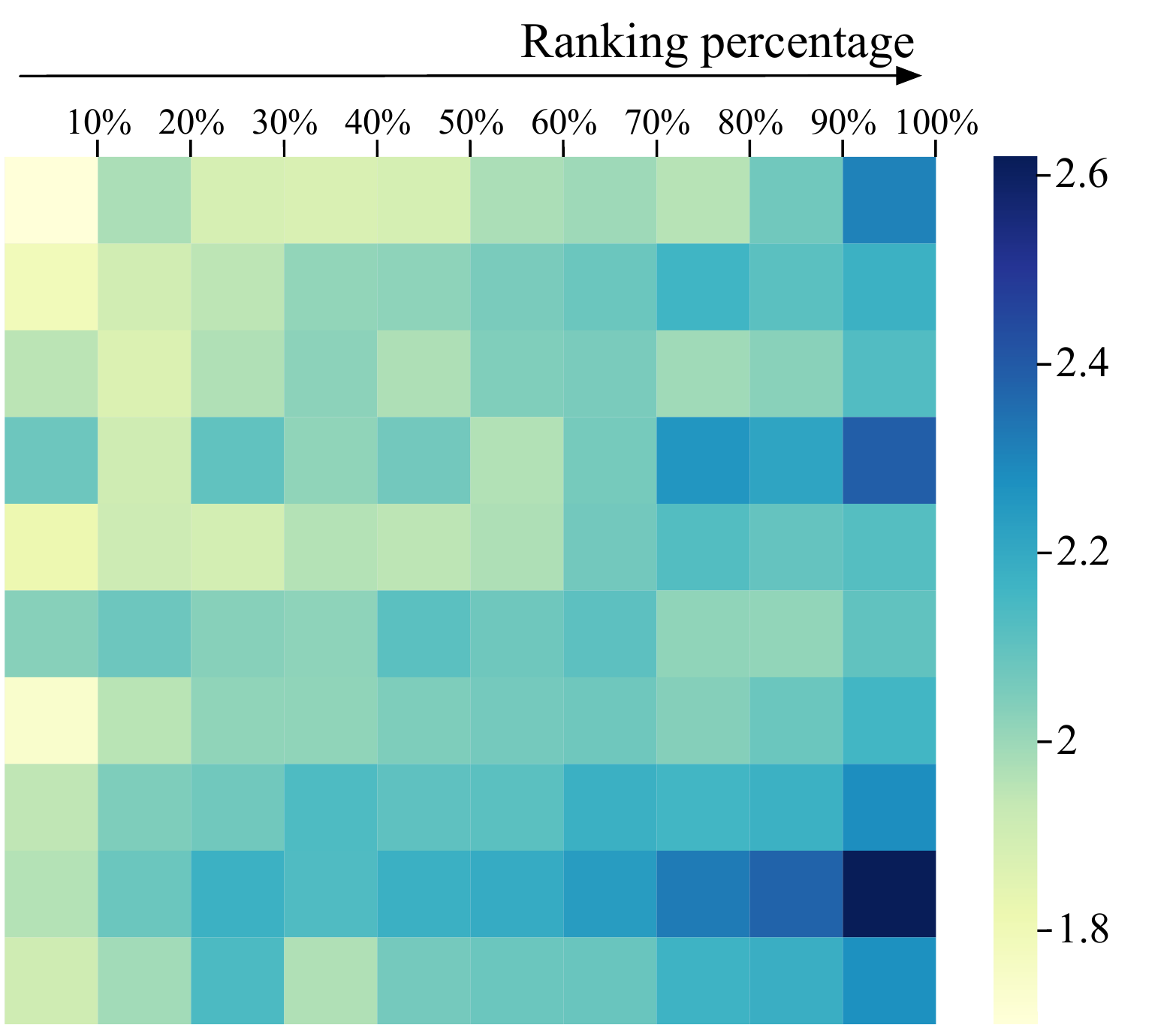}
    }
    \hspace{3mm}
    \subfloat[Heatmap: geographical \\distance between counties.]
    {
    	\label{pear-b}
        \includegraphics[width=0.212\textwidth]{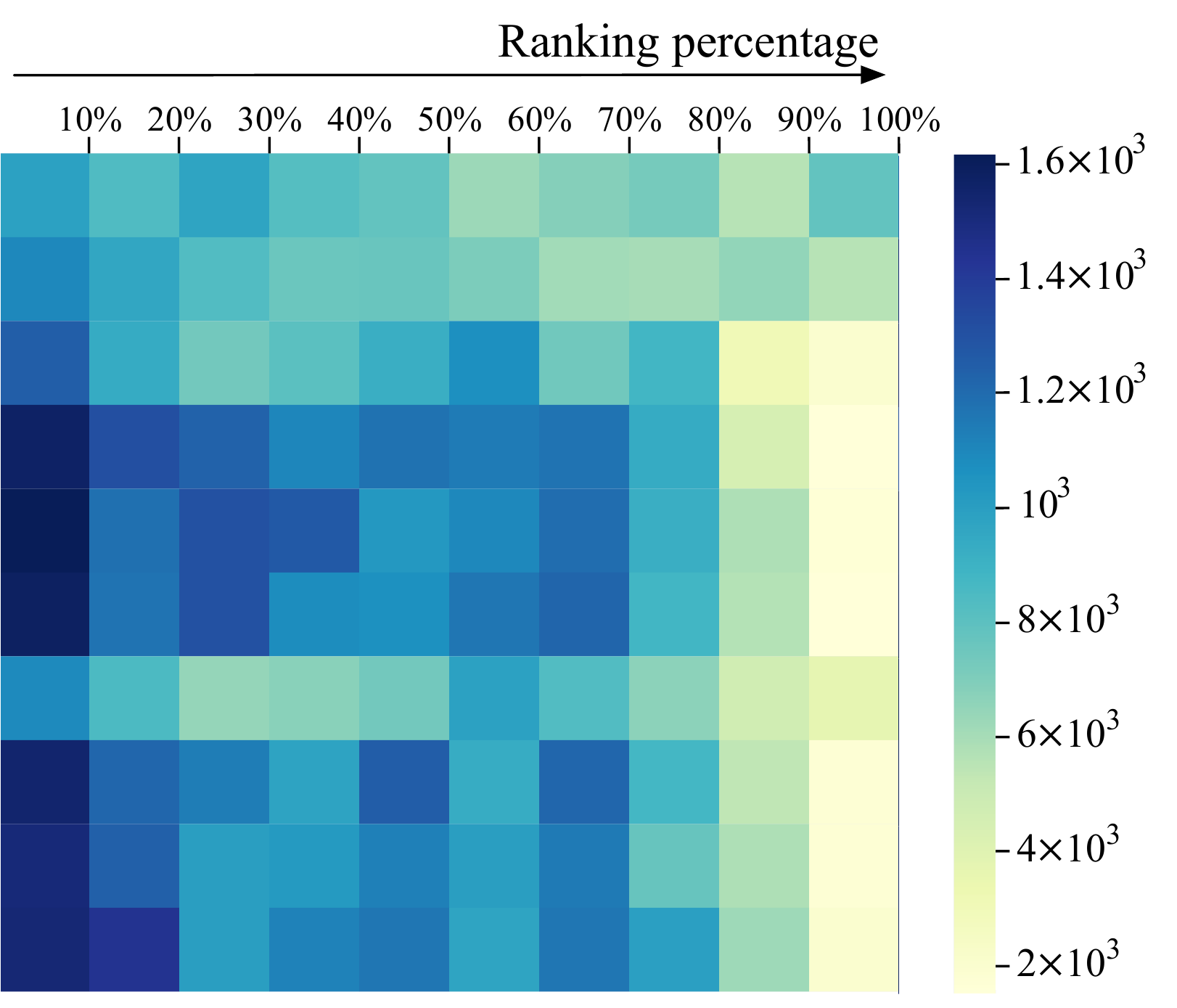}
    }
    \subfloat[Heatmap: mobility flow \\volume between counties.]
    {
    	\label{pear-c}
        \includegraphics[width=0.212\textwidth]{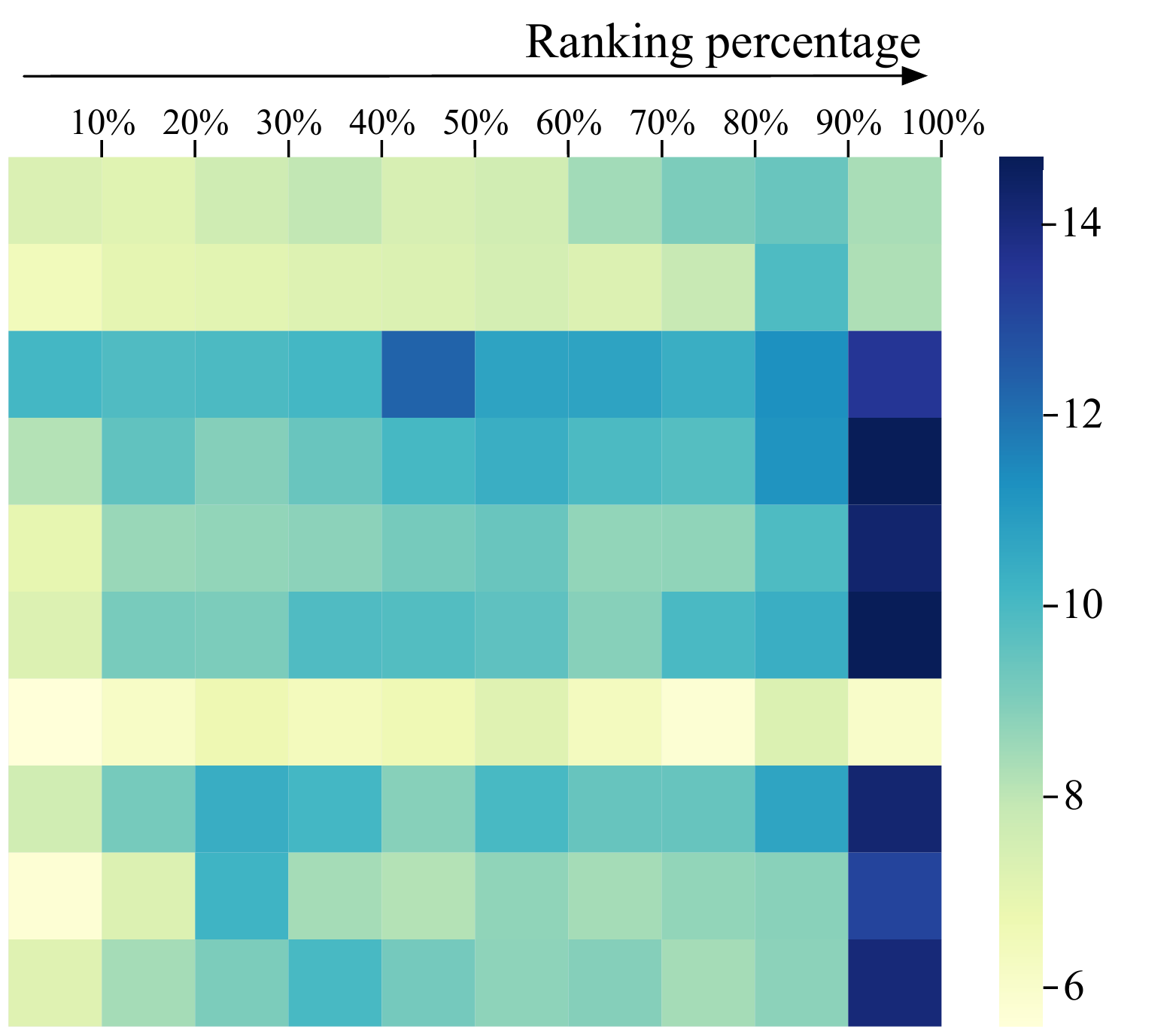}
    }
	\caption{Illustrations reflecting interactions between the selected counties and other counties: 
	(a) Pearson correlation of the number of confirmed cases, (b) Pearson correlation of keyword popularity, (c) geographic distance, and (d) mobility flow volume between counties. The counties in each row of (a) are ranked by the Pearson correlation of the confirmed case number in an ascending order, and all the results are averaged over every 10 percentile of counties. Each row in (b), (c) and (d) follows the same order of the counties as in (a). 
}  
	\label{pearson}
\end{figure*}

\vspace{-0.1in}
\subsection{Preliminary Data Analysis}

To explore whether the covariates and networks have the potential to capture the (unobserved) confounders, we conduct preliminary data analysis to explore their dependencies with COVID-19 outbreak dynamics (i.e., the number of confirmed/death cases). Due to the space limit, we only show the analysis on the cumulative confirmed cases number of ten counties from Virginia (VA) and Massachusetts (MA) as examples. Similar observations can also be found in other counties, as well as the number of death cases. 

\noindent\textbf{Relations between covariates and outbreak dynamics.} As proxy variables of unobserved confounders, covariates of counties could have dependencies with the COVID-19 outbreak dynamics (outcome). For example, counties with relatively higher similarity of covariates may also have higher similarity in the number of confirmed cases. If such dependencies exist, it implies that the covariates of counties could be potentially helpful in capturing the unobserved confounders. In this regard, we explore whether such dependency exists in our collected covariates of counties, i.e.,  popularity of COVID-19 related search keywords of counties on Google US. We first compute the Pearson correlation of the daily cumulative confirmed case number series in 2020 between the chosen counties and all 391 counties. Then for each of the ten counties, we rank its Pearson correlation values with the $391$ counties in an ascending order. The average Pearson correlation value over every 10 percentile of the ranking is reported in Fig. (\ref{pear-a}). 
Due to space constraints, we explain the process here only for one of the 19 keywords we used, "social distance".
Similar observations can also be drawn based on other keywords. For each county, we represent the daily popularity of "social distance" on Google US as a time series spanning from February to December 2020. The Pearson correlation between counties based on their daily popularity is then calculated. Following the same ranking order in each row of Fig. (\ref{pear-a}), we report the average Pearson correlation value of their daily popularity over every 10 percentile in Fig. (\ref{pear-d}) in exponential scale. The results implies that most county pairs with higher Pearson correlation w.r.t. outbreak dynamics tend to have higher keyword popularity similarity. This reveals that our collected covariates have the potential to help capture the confounders.

\noindent\textbf{Relations between networks and outbreak dynamics.} Networks could also have dependencies with COVID-19 outbreak dynamics. For example, 
county pairs with relatively shorter distance or larger mobility flow volume may also have higher similarity in the number of confirmed cases. 
Similar to the 
relationship
between covariates and outbreak dynamics, such dependencies imply the potential utility of networks in capturing the confounders. Consequently, here we explore whether such dependencies exist in networks among counties. Following the same ranking order in each row of Fig. (\ref{pear-a}), we also report the corresponding value of their distance and mobility flow (aggregated between Feb. and Dec. in 2020) averaged over every 10 percentile (in log scale) in Fig. (\ref{pear-b}) and Fig. (\ref{pear-c}), respectively. The following conclusions can be drawn: firstly, most county pairs with relatively lower Pearson correlation w.r.t. the outbreak dynamics are more likely to have larger distance than those with high correlation; secondly, most county pairs with relatively higher Pearson correlation w.r.t. the outbreak dynamics tend to have larger human mobility flow volume between them. 
This implies
that our collected networks have the potential to capture the confounders.

%% file: method.tex
\section{Time-varying Causal Assessment}
In this section, we formulate the causal assessment of COVID-19 related policy types as a causal effect estimation problem from time-varying observational data. To mitigate the confounding bias of such assessment, we develop a neural network based framework to capture the time-varying confounders.


\vspace{-0.1in}
\subsection{Formulating Policy Assessment as A Causal Effect Estimation Problem}

We consider the COVID-19 related policy types in $n$ counties across $T$ time periods. 
Different counties are described by the same set of covariates (a.k.a. features) over time, and we denote them by  $\bm{X}^t=\{\bm{x}_1^t,...,\bm{x}_{n}^t\}$, where $\bm{x}_i^t$ represents the covariates of the $i$-th county at time period $t$ (e.g., in 
Albemarle county, VA, residents' search popularity of COVID-19 related keywords on Google throughout March, 2020). We represent the adjacency matrix of the network (e.g., the distance network or the mobility flow network) at time period $t$ as $\bm{A}^t\in \mathbb{R}^{n\times n}$, where $\bm{A}^t_{ij}$ is the weight of edge $i\rightarrow j$ in $\bm{A}^t$, and $\bm{A}^t_{ij}=0$ if there does not exist such edge. We assume that the edge weight can reflect the similarity between counties. Intuitively, the larger the weight is, the more similar these two counties are.
For each policy type, we use $\bm{C}^t=\{c_1^t,...,c_{n}^t\}$ to denote whether policies of this type are in effect in these $n$ counties at time period $t$, where $c_i^t$ is either $1$ (treated) or $0$ (not treated, a.k.a. controlled), corresponding to whether the policy type is in effect in the $i$-th county or not. At each time period, the treated counties form the \textit{treated group}, while the controlled counties form the \textit{control group}.
Here, we denote a specific manifestation of outbreak dynamics (such as the number of confirmed cases) at time period $t$ as $\bm{Y}^t=\{y_{1}^t,...,y_{n}^t\}$, which are also referred to as \textit{observed outcomes}. The superscript``$<t$'' represents the historical data before time period $t$, e.g., the covariates before time period $t$ are $\bm{X}^{<t}=\{\bm{X}^1,\bm{X}^2,...,\bm{X}^{t-1}\}$, and  $\bm{C}^{<t}, \bm{A}^{<t}$ are defined similarly. Additionally, we use $\mathcal{\bm{H}}^{t}=\{\bm{X}^{<t},\bm{A}^{<t},\bm{C}^{<t},\bm{Y}^{<t}\}$ to denote all the historical observational data before time period $t$.

To assess the impact of COVID-19 related policies from a causal perspective, we frame this assessment as a causal effect estimation problem from time-varying observational data. The goal of this problem is to investigate to what extent a \textit{cause} (a.k.a. \textit{treatment}, e.g., a policy in effect) would causally affect an \textit{outcome} (e.g., the number of confirmed cases) for each instance (e.g., a county) at different time periods.
To estimate the causal effect of a policy on the outbreak dynamics at time period $t$, we should compare the potential outcomes of the outbreak dynamics in each county if this policy had/had not been in effect during time period $t$.
Generally, the \emph{potential outcome} \cite{neyman1923applications,rubin2005bayesian} means the outcome that would be realized if the instance got treated/controlled, e.g., ``In March, what would the number of confirmed cases be in Albemarle county, VA, if the mask requirement policy had/had not been in effect during that time period?". 
We represent the potential outcomes of all counties if the policy had/had not been in effect at time period $t$ by $\bm{Y}_1^t=\{y_{1,1}^t,...,y_{1,n}^t\}$ and $\bm{Y}_0^t=\{y_{0,1}^t,...,y_{0,n}^t\}$. 
In our setting, the potential outcome $y_{c,i}^t$ is the outcome that would be realized if the $i$-th instance is under treatment $c$ at time period $t$. 
Then the \textit{individual treatment effect (ITE)} \cite{rubin2005bayesian} for each instance at time period $t$ is defined as the difference between the potential outcome if the instance gets treated and the potential outcome if it gets controlled at that time period. In our problem, the ITE of each policy on the outbreak dynamics in each county at time period $t$ is the difference between two potential outcomes:
\begin{equation}
    \tau_i^t = \mathbb{E}[y_{1,i}^t|\bm{x}_i^{t}, \bm{A}^{t},\mathcal{\bm{H}}^{t}] -\mathbb{E}[y_{0,i}^t|\bm{x}_i^{t}, \bm{A}^{t},\mathcal{\bm{H}}^{t}].
    \label{eq:ite}
\end{equation}
Then the average treatment effect (ATE) at time period $t$ is computed as the average of ITEs over all counties:
\begin{equation}
    \tau_{ATE}^t=\frac{1}{n}\sum_{i=1}^{n}\tau_i^t.
\end{equation}

\begin{figure}[!t]
	\centering  
	\includegraphics[width=0.9\linewidth,height=1.75in]{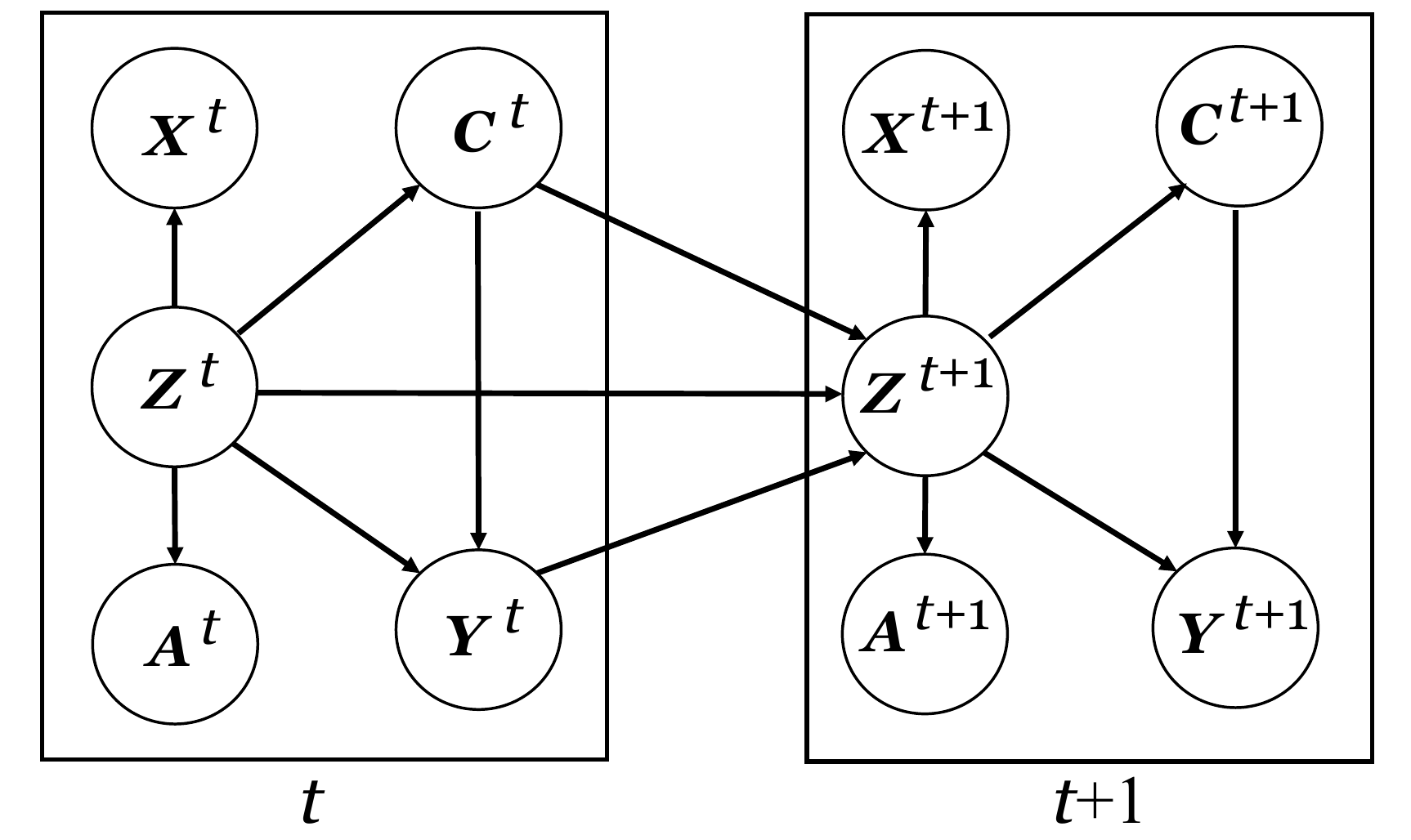}
	\vspace{-0.3cm}
	\caption{Causal graph of our studied problem.} 
	\label{fig:causal_graph}  
	\vspace{-0.2cm}
\end{figure}

Motivated by our previous work on causal effect estimation from time-evolving data~\cite{ma2021deconfounding}, we design a causal graph for our studied problem as shown in Fig.~\ref{fig:causal_graph}, where each arrow means a causal relationship. 
We denote the time-varying (unobserved) confounders  (e.g., residents' vigilance) by $\bm{Z}^t=\{\bm{z}_1^t,...,\bm{z}_{n}^t\}$. 
To achieve unbiased treatment effect estimation, we need to capture these confounders. 
Specifically, we use a weaker version of the unconfoundedness assumption \cite{rubin2005bayesian} by assuming that there may exist unobserved confounders, and conditioning on all the confounders, the treatment assignment is independent with the potential outcomes, i.e., $y_{1,i}^t,y_{0,i}^t \bigCI c_i^t| \bm{z}_i^t$. With this assumption, the unbiased treatment effect estimation can be obtained if we capture all the (unobserved) confounders from the observational data, as proved in \cite{ma2021deconfounding}.

\subsection{Estimating Causal Effects of COVID-19 Related Policies on Outbreak Dynamics}
To better estimate the causal effects of COVID-19 related policies on the outbreak dynamics at each time period, we develop a neural network based framework with input as the time-varying observational data. Specifically, the proposed framework aims to learn representations of time-varying confounders by predicting the COVID-19 outbreak dynamics and whether the policies are in effect at each time period.


\subsubsection{Learning Representations for Time-varying Confounders}
Our framework captures the time-varying confounders from the evolving observational data at different time periods.
As shown in Fig.~\ref{fig:causal_graph}, we assume the confounders causally affect both the treatment assignment and the outcome at the current time period, while the treatment assignments, outcomes and confounders at the previous time period also affect the current confounders. To capture the confounders from the time-varying observational data, we first use recurrent neural networks (RNNs) \cite{medsker2001recurrent} to extract useful historical information from the data at previous time periods: 
\begin{equation}
\vspace{-1mm}
     \bm{H}^{t} = \text{RNN}({\bm{H}}^{t-1}, (\bm{Z}^t \oplus \bm{X}^t \oplus \bm{C}^t \oplus \bm{Y}^t)),
\end{equation}
where $\bm{H}^t$ is the hidden unit in the RNNs which captures the historical information, including the previous treatment assignment $\bm{C}^{<t}$ (i.e., whether a policy was in effect), outcomes $\bm{Y}^{<t}$ (i.e., numbers of confirmed or death cases), network structure $\bm{A}^{<t}$ (e.g., distance network) and observed features $\bm{X}^{<t}$ (e.g., search popularity of COVID-19 related keywords on Google). Here $\oplus$ denotes a concatenation operation. 
Besides, to better capture the unobserved confounders, we also take advantage of the network structure among counties using 
graph convolutional networks (GCNs) \cite{kipf2016gcn}. Specifically, at each time period, we learn the representation of current confounders from the hidden state $\bm{H}^{t-1}$, as well as the current network structure $\bm{A}^t$ and covariates $\bm{X}^t$:
\begin{equation}
\bm{Z}^t =\hat{\bm{A}}^t \text{ReLU}(\hat{\bm{A}}^t (\bm{X}^t\oplus{\bm{H}}^{t-1}) \bm{W}_0)\bm{W}_1,
\end{equation}
where we stack two GCN layers to capture the non-linear dependencies between the unobserved confounders and the input data. $\bm{W}_0$, $\bm{W}_1$ are the parameters of the GCNs layers. $\hat{\bm{A}}^t$ is the normalized adjacency matrix computed from $\bm{A}^t$ beforehand with the renormalization trick \cite{kipf2016gcn}. Specifically,
$\hat{\bm{A}}^t=(\tilde{\bm{D}}^t)^{-\frac{1}{2}}\tilde{\bm{A}}^t(\tilde{\bm{D}}^t)^{-\frac{1}{2}}$, where $\tilde{\bm{A}}^t=\bm{A}^t+\bm{I}_n$, $\tilde{\bm{D}}^t_{jj}=\sum_j \tilde{\bm{A}}_{jj}$. 
In this way, we establish a representation learning module for the time-varying confounders.

\noindent\textbf{Balancing the Representations of Confounders.} 
For certain COVID-19 related policies, the counties that had this policy and those that did not often have different distributions of confounders, i.e., the distributions of confounders in the treated and control groups are very imbalanced. Therefore, we employ representation balancing techniques \cite{CFR} at each time period by adding a distribution balancing constraint $\mathscr{L}_b$, which is the Wasserstein-1 distance \cite{CFR} of the representation distributions between the treated group and control group. Minimizing the constraint can encourage the representation distributions of these two groups to be closer, and has been proved to benefit the causal effect estimation~\cite{CFR}.

\subsubsection{Predicting Outbreak Dynamics and Presence of Policies with Confounder Representations}

To train the model to capture the representations of time-varying confounders, we take advantage of the observed outcome (i.e., outbreak dynamics) and treatment assignment (i.e., whether a policy is in effect) as supervision signals. Firstly, we use two multilayer perceptrons (MLPs) $f_1$ and $f_0$ to model the potential outcomes, and denote the predicted outcomes as:
\begin{equation}
    \hat{y}_{1,i}^t=f(\bm{z}_i^t, c_i^t=1)= f_1(\bm{z}_i^t),
\end{equation}
\begin{equation}
    \hat{y}_{0,i}^t=f(\bm{z}_i^t, c_i^t=0) = f_0(\bm{z}_i^t).
\end{equation} 
This way, each instance's \textit{factual outcome} $y_{F,i}^t$ (observed outcome, e.g., the observed number of confirmed cases) and \textit{counterfactual outcome} $y_{CF,i}^t$ (unobserved outcome with the contrary treatment, e.g., the potential number of confirmed cases 
if the policy had been treated differently from the fact) are estimated. The mean squared error (MSE) is used as the factual outcome prediction loss:
\begin{equation}
    \mathscr{L}_y = \mathbb{E}_{t\in [T],i\in[n]}[(\hat{y}_{F,i}^t-y_{F,i}^t)^2].
\end{equation} Secondly, we also utilize the treatment assignment to train the representations of confounders. Specifically, we use a binary prediction module activated by a softmax layer. The output logit $\hat{s}_i^t$ estimates the probability of getting treated, i.e., whether a policy is in effect in each county at time period $t$. The cross-entropy loss is adopted for the treatment assignment prediction:
\begin{equation}
    \mathscr{L}_c=-\mathbb{E}_{t\in [T],i\in[n]}[ (c_i^t\log(\hat{s}_i^t)+(1-c_i^t)\log(1-\hat{s}_i^t))].
\end{equation}
The overall loss function of our framework is:
\begin{equation}
    \mathscr{L}=\mathscr{L}_y + \alpha\mathscr{L}_c + \beta\mathscr{L}_b + \lambda\|\bm{\Theta}\|^2,
\end{equation}
where $\alpha,\beta,\lambda$ are positive balancing hyperparameters, and $\Theta$ denotes the set of model parameters to train. After training the model, we estimate the ITE of the studied COVID-19 policy in each county $i$ at time period $t$ as the difference between the predicted potential outcomes $\hat{\tau}_i^t = \hat{y}_{1,i}^t - \hat{y}_{0,i}^t$ and assess the ATE by averaging the estimated ITEs over all counties.

%% file: exp.tex
\vspace{-2mm}
\section{Experimental Evaluations}
In this section, we first show the estimated causal effects of different policies on the outbreak dynamics 
(number of confirmed cases and death cases) at different time periods during 2020. Based on the estimation results, we assess the impact of these policies both on a macro-level and micro-level, and discuss our assessment with regard to the  details of the corresponding policies.
Then we evaluate our framework with respect to its ability of controlling for confounding. Due to the lack of counterfactual data, we compare our framework with other state-of-the-art causal effect estimation methods in the performance of predicting the outbreak dynamics (factual outcome prediction) and predicting whether the policies are in effect (treatment prediction). We also compare the causal effects estimated by different methods for further discussion.

\begin{table*}
\vspace{0.2cm}
\centering
\caption{Information of selected policies in each category.} 
\label{tab:policy}
\small
\begin{tabular}{l|l|l}
\hline
Categories                                                                           & Top policy types &  
Example policies (including the states that enacted them)              \\
\hline
\multirow{5}{*}{\begin{tabular}[c]{@{}l@{}} Social Distancing \\ (Keywords:``gather'', etc.)\end{tabular}}  & State of Emergency & Limit in-person meetings and non-essential, work-related gatherings (VA)                        \\  
                                                                                                                   & Nursing homes &Limit or avoid unnecessary visits to hospitals, nursing homes, and residential care facilities (OH)                                \\
                                                                                                                      & Food and Drink  & Outdoor dining, take-out, and delivery only (RI); All bars will close, no in-house dining (VI)                       \\
                                                                                                                      & Childcare (K-12)& Close all public schools (VI); Public/private schools are asked to delay in-person instruction (KY) \\
                                                                                                                      & Gatherings & Mass gatherings remain limited to 10 people (OH); 100 people outdoors, 10 people indoors (MI)                                  \\
\hline
\multirow{5}{*}{\begin{tabular}[c]{@{}l@{}} Reopening \\ (Keywords:``reopen'', etc.)\end{tabular}}                 & Phase 2 & Allow some lodging establishments that have completed a specific training to reopen (NM)                  \\
                                                                                                                      & Entertainment & Businesses, move theaters, health and athletic clubs, may reopen (ND)                             \\
                                                                                                                      & Outdoor and recreation& Contact sport practices and non-contact sport competitions to reopen (OH)                    \\
                                                                                                                      & Personal care & Self-care may reopen (ND); massage therapy may reopen to customers with appointments (AR)                          \\
                                                                                                                      & Food and drink & Reopen on-site dining  (TN); restaurants can reopen for limited services under guidelines (SC)                           \\
\hline
\multirow{5}{*}{\begin{tabular}[c]{@{}l@{}} Mask Requirement \\ (Keywords:``mask'', etc.)\end{tabular}}            & Any mask Requirement & Face covering required in businesses, public buildings, indoor public spaces, transportation (IN)                           \\
                                                                                                                      & Public facing bussinesses &  Mandate face mask use by all individuals in public facing bussinesses (WI)                      \\
                                                                                                                      & Food and drink  & Employees and patrons in restaurants are required to wear facial coverings (MP)                           \\
                                                                                                                      & Phase 3 & Outdoor venues must follow guidelines including mask wearing (ID)\\
                                                                                                                      & Phase 2& Public Health guidelines include mask wearing (ID); mandating masks in all K-12 schools (UT)\\
\hline
\end{tabular}
\end{table*}

\noindent\textbf{Setup.} 
Counties are randomly assigned to
training data ($60\%$), validation data ($20\%$), and test data ($20\%$). 
The reported results 
are the average of $10$ repeated executions. Unless otherwise 
stated,
we set the length of each time period as $15$ days, 
the
learning rate as $3e-3$, epochs as $2,000$, the dimension size of 
hidden states $d_h=50$, the dimension size of the confounder representation $d_z=50$, $\alpha=1.0$, $\beta=1e-4$, $\lambda=0.01$, and we use an Adam optimizer. 


\subsection{Causal Assessment of COVID-19 Related Policies on Outbreak Dynamics}
\begin{figure*}[!t]
	\centering 
	\subfloat[Social distancing, confirmed cases.]{
	\label{effect_confirm_SD}
        \includegraphics[width=0.33\textwidth,height=1.2in]{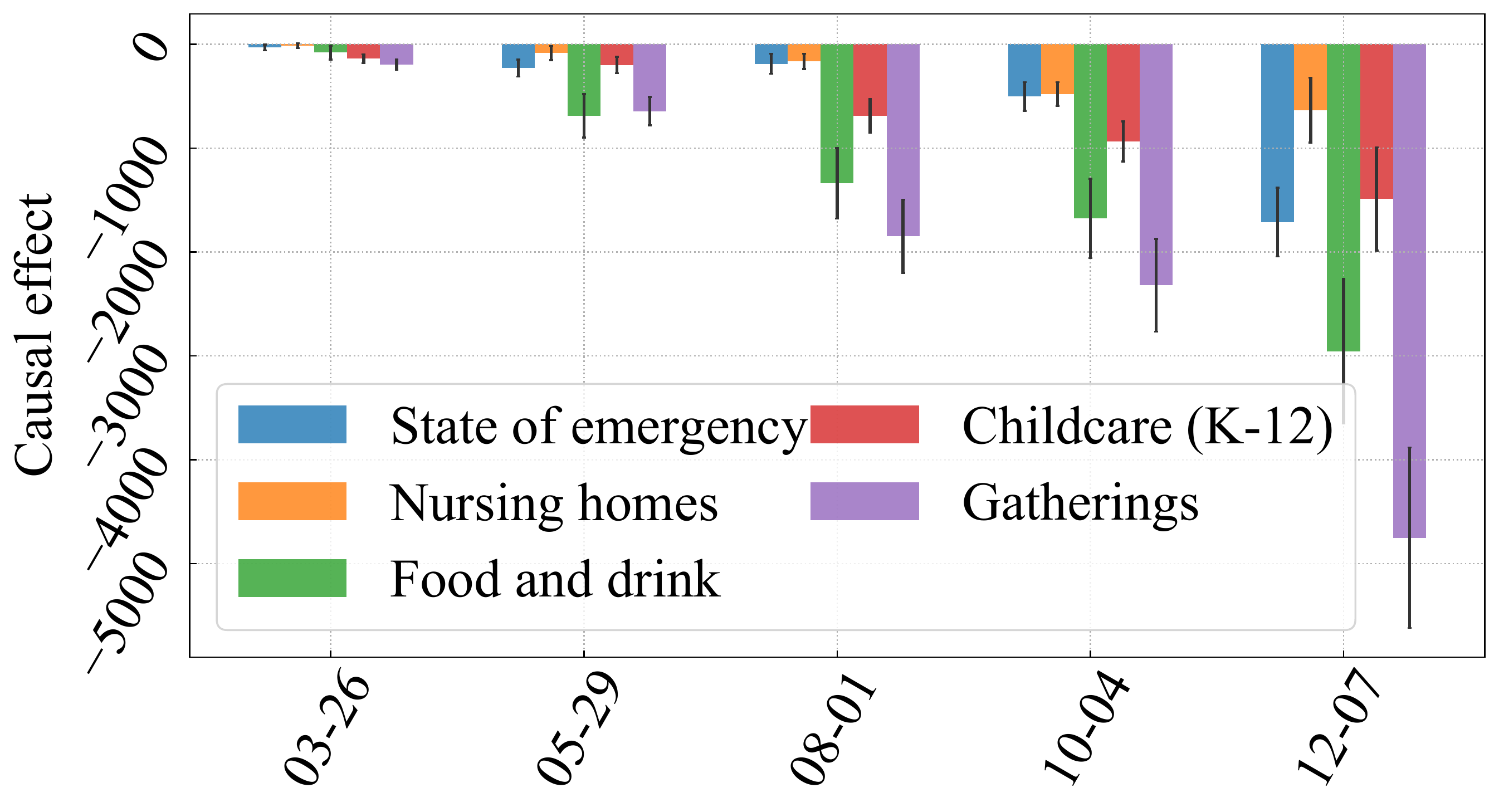}
    } 
    \subfloat[Reopening, confirmed cases.]{
    	\label{effect_death_SD}
        \includegraphics[width=0.33\textwidth,height=1.2in]{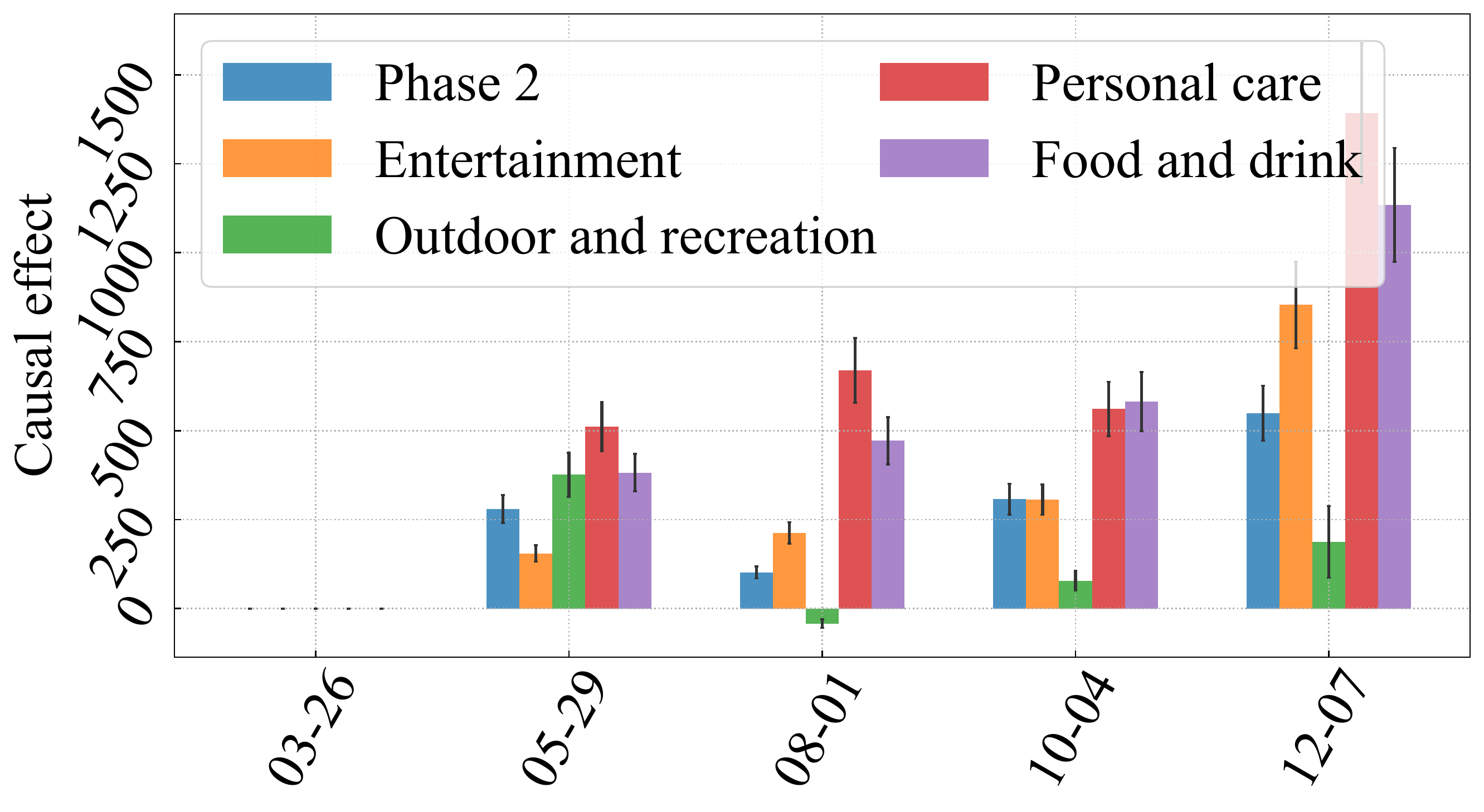}
    }
    \subfloat[Mask requirement, confirmed cases.]{
    	\label{assign_SD}
        \includegraphics[width=0.32\textwidth,height=1.2in]{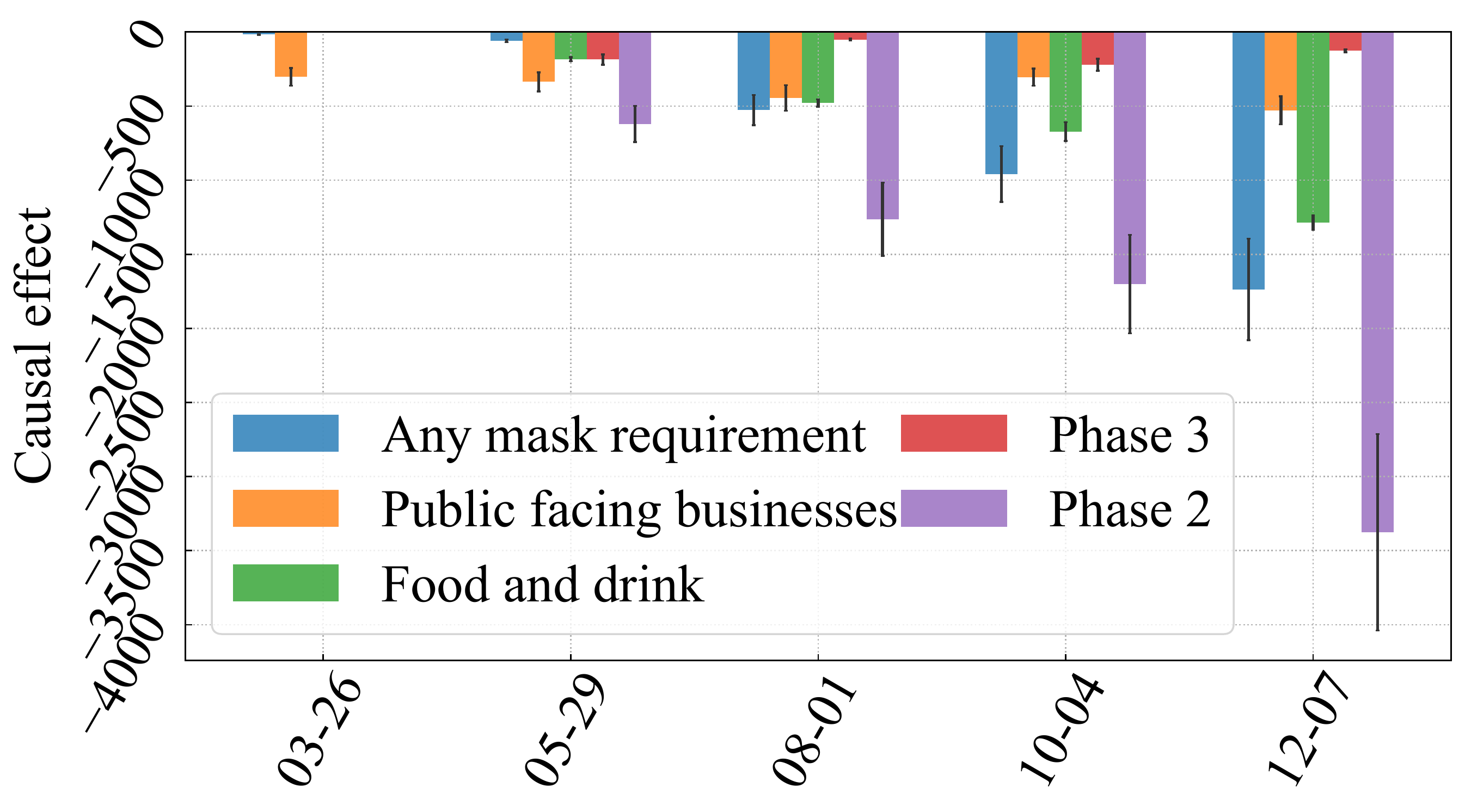}
    }
    
    \subfloat[Social distancing, death cases.]{
	\label{effect_confirm_RO}
        \includegraphics[width=0.33\textwidth,height=1.2in]{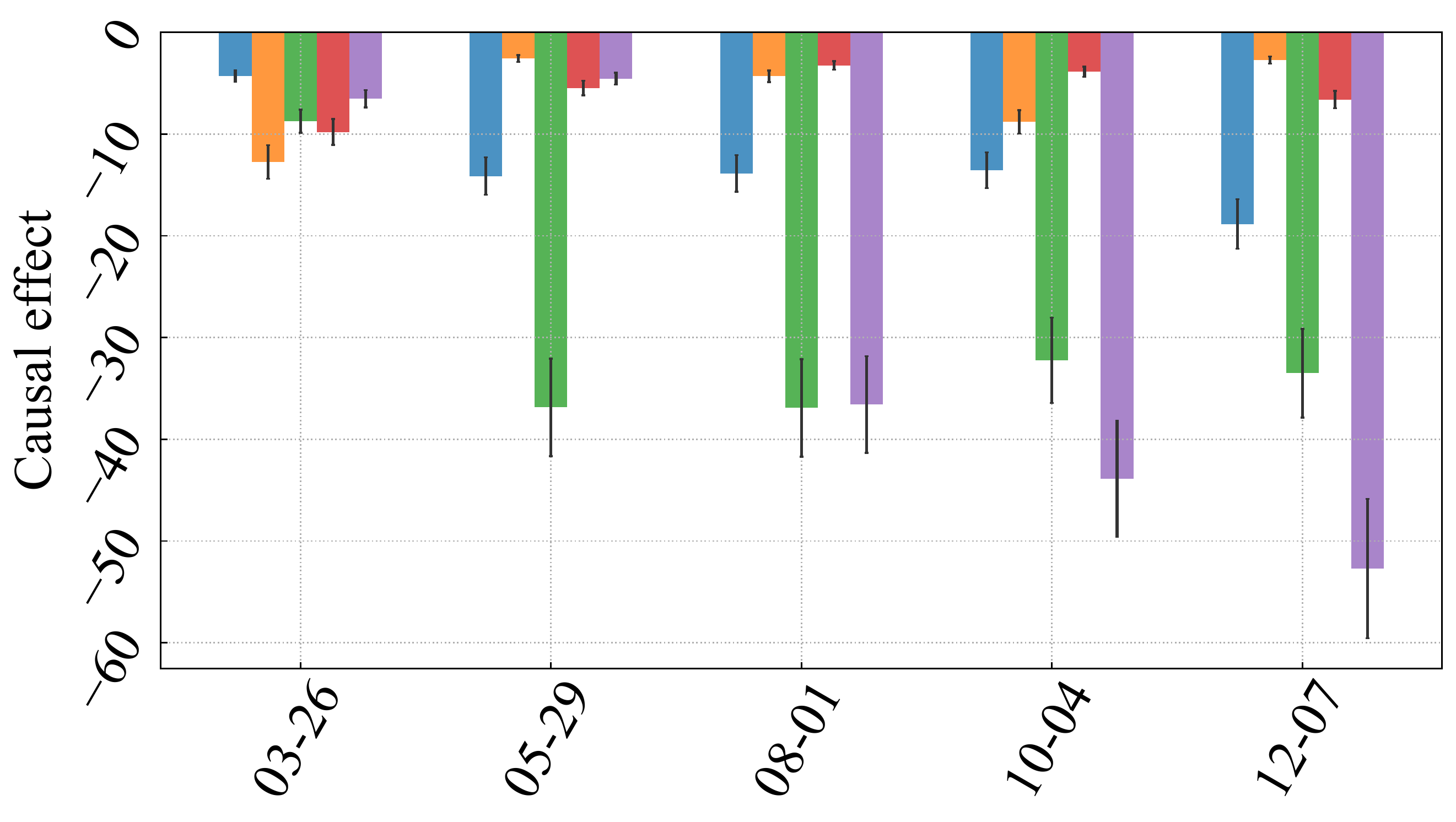}
    } 
    \subfloat[Reopening, death cases.]{
    	\label{effect_death_RO}
        \includegraphics[width=0.33\textwidth,height=1.2in]{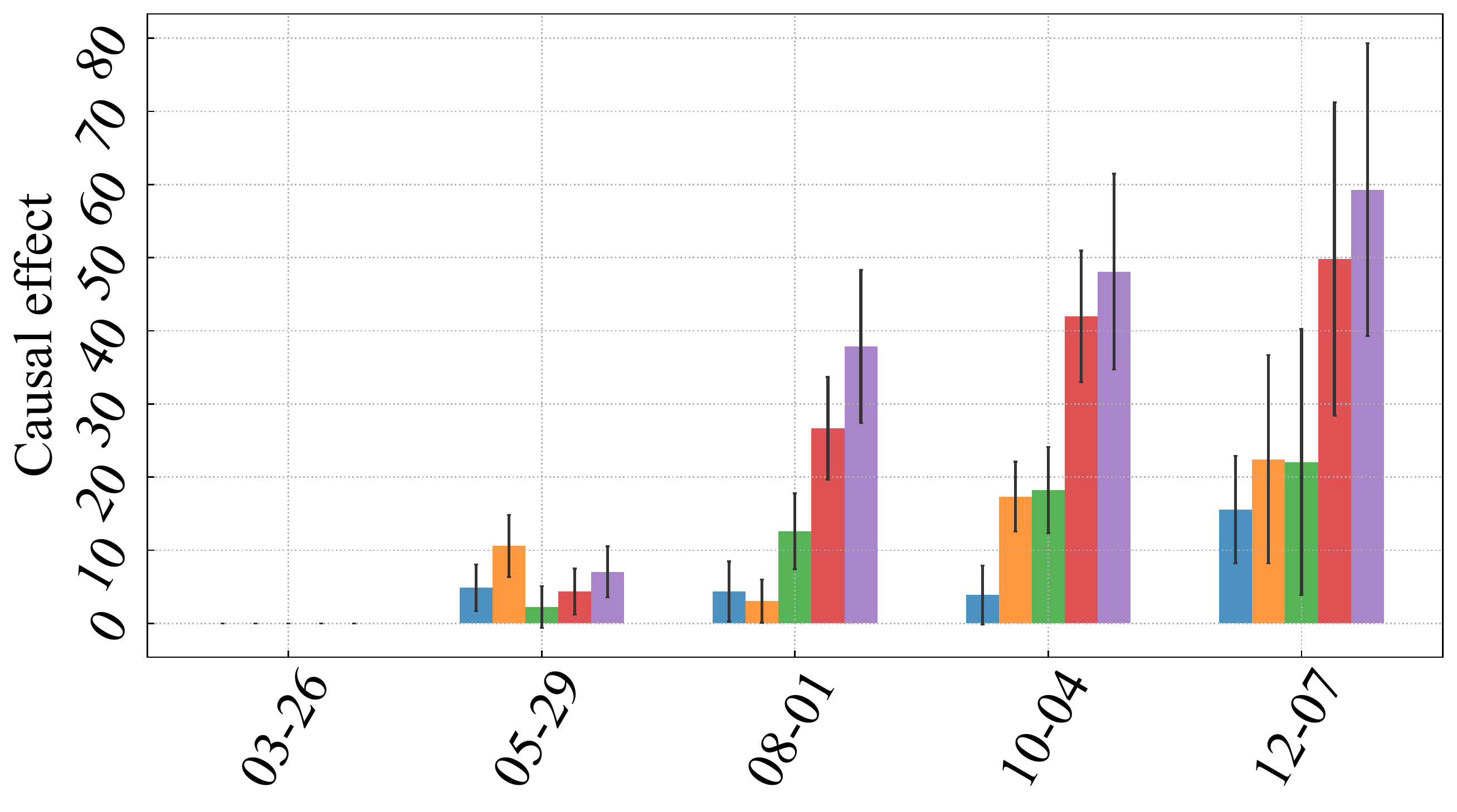}
    }
    \subfloat[Mask requirement, death case.]{
    	\label{assign_RO}
        \includegraphics[width=0.32\textwidth,height=1.2in]{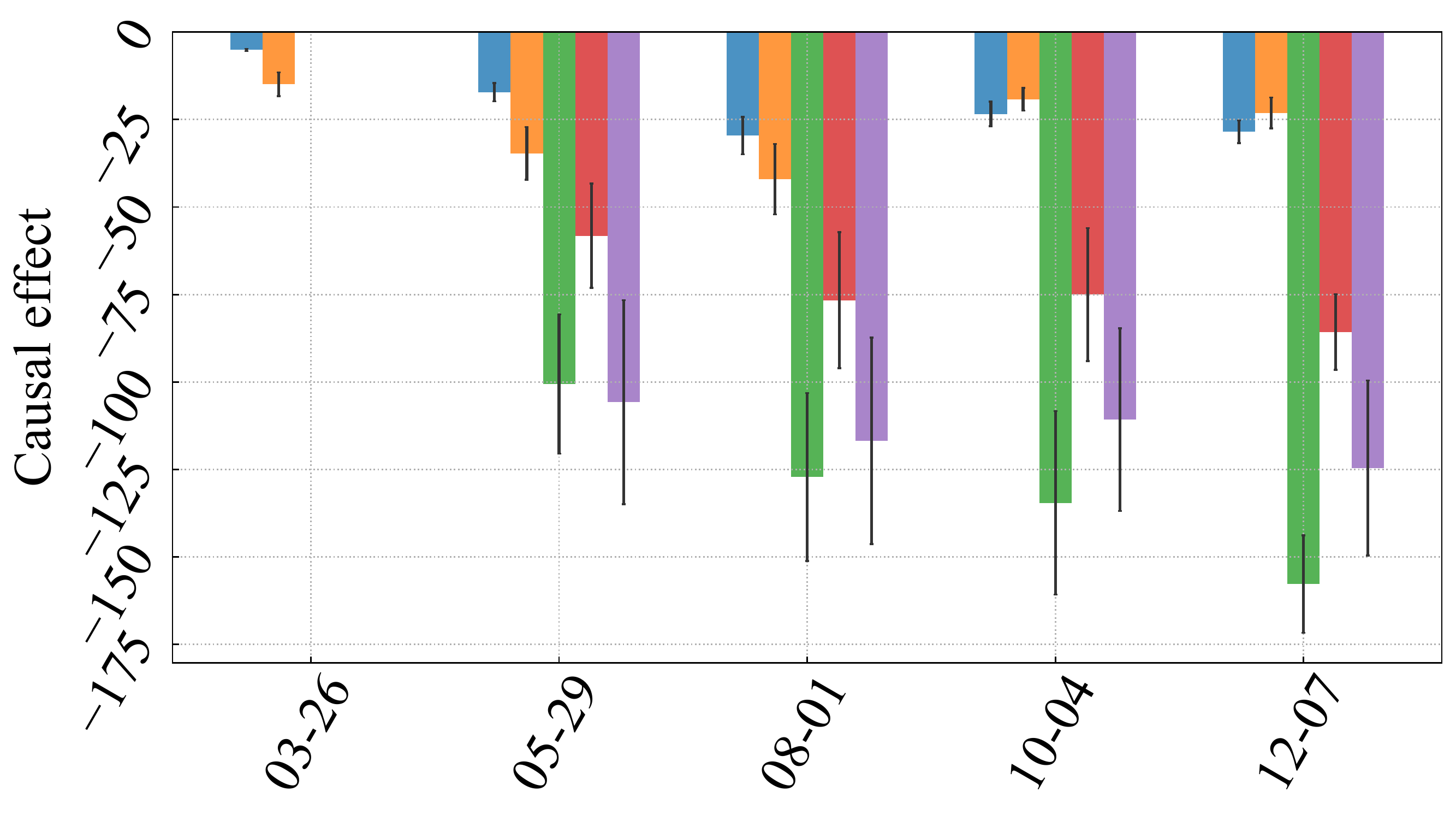}
    }
	\caption{Causal effect estimation of different policies over time. The three columns correspond to the policy categories of social distancing, reopening, and mask requirements. 
	Each panel shows data for several policy types.
    The first row is the estimated causal effect of the respective policy type {\textit {on the number of confirmed cases}}; the second row 
	{\textit {on the number of deaths.}}
	}
	\label{fig:causal_effect}
\end{figure*}

We estimate the causal effects of different policies on the outbreak dynamics at different time periods during 2020. 
For each category of policies with a certain goal, we select the policy types adopted by over $10\%$ of counties, and Table~\ref{tab:policy} shows the information about the top-$5$ most impactful policy types in each category based on our estimation of their average treatment effects (ATEs). 
Fig.~\ref{fig:causal_effect} summarizes our estimation of the ATEs of these policy types in each category.
Each column in Fig.~\ref{fig:causal_effect} corresponds to a category of policies: \emph{social distancing}, \emph{reopening} and \emph{mask requirement}. The first and second rows show the estimated ATEs of these policy types on the number of confirmed cases, and the number of death cases, respectively. 
We have the following observations from Fig.~\ref{fig:causal_effect}:
\begin{itemize}
\item At the macro-level, the policy types regarding social distancing and mask requirement have negative causal effects on both the number of confirmed cases and death cases, while the policy types about reopening have positive causal effects. These negative values of causal effect indicate that the corresponding policy types (e.g. those in the social distancing and mask requirement categories) causally help reduce the spread of COVID-19, while the policy types with positive values of causal effects (e.g., those in the reopening category) may have a contrary effect because they increase the risk of infection. These observations 
appear intuitively plausible
and are also consistent with existing literature regarding COVID-19 related policies \cite{hsiang2020effect,kristjanpoller2021causal,mitze2020face}. 
\item At the micro-level, we zoom into the most impactful policy types in each category. In the category of social distancing, the policy types ``Gatherings'' and ``Food and drink'' seem to have the strongest effects. From the description of the detailed policies, they powerfully prohibit the number of individuals in multiple activities, especially in high-risk places such as indoor restaurants. In the category of reopening, the estimated effects of policy types ``Personal care'' and ``Food and Drink'' indicate that reopening public places such as personal care center and restaurants heavily increase the risk of COVID-19 infection. In the category of mask requirement, the policy types ``Phrase 2'', ``Any mask requirement'', and ``Food and drink'' are the most impactful as they mandate face masks used by individuals in many different public spaces.
\item Generally, above observations are consistent for different outcomes including the number of confirmed cases and death cases, as well as for different time periods. Besides, we observe that the policies can have stronger effects when the the pandemic becomes more severe, such as during the outbreak at the end of 2020. In conclusion, above observations reveal the importance of in-time policies to limit close contact (within about 6 feet) among people in different aspects, e.g., distance, frequency and certain body parts (e.g., face) during the spread of respiratory pandemics like COVID-19.
\end{itemize}



\begin{figure*}[!t]
	\centering 
	\vspace{-1mm}
	\subfloat[Confirmed cases.]{
	\label{ite_confirm}
        \includegraphics[width=0.47\textwidth,height=1.5in]{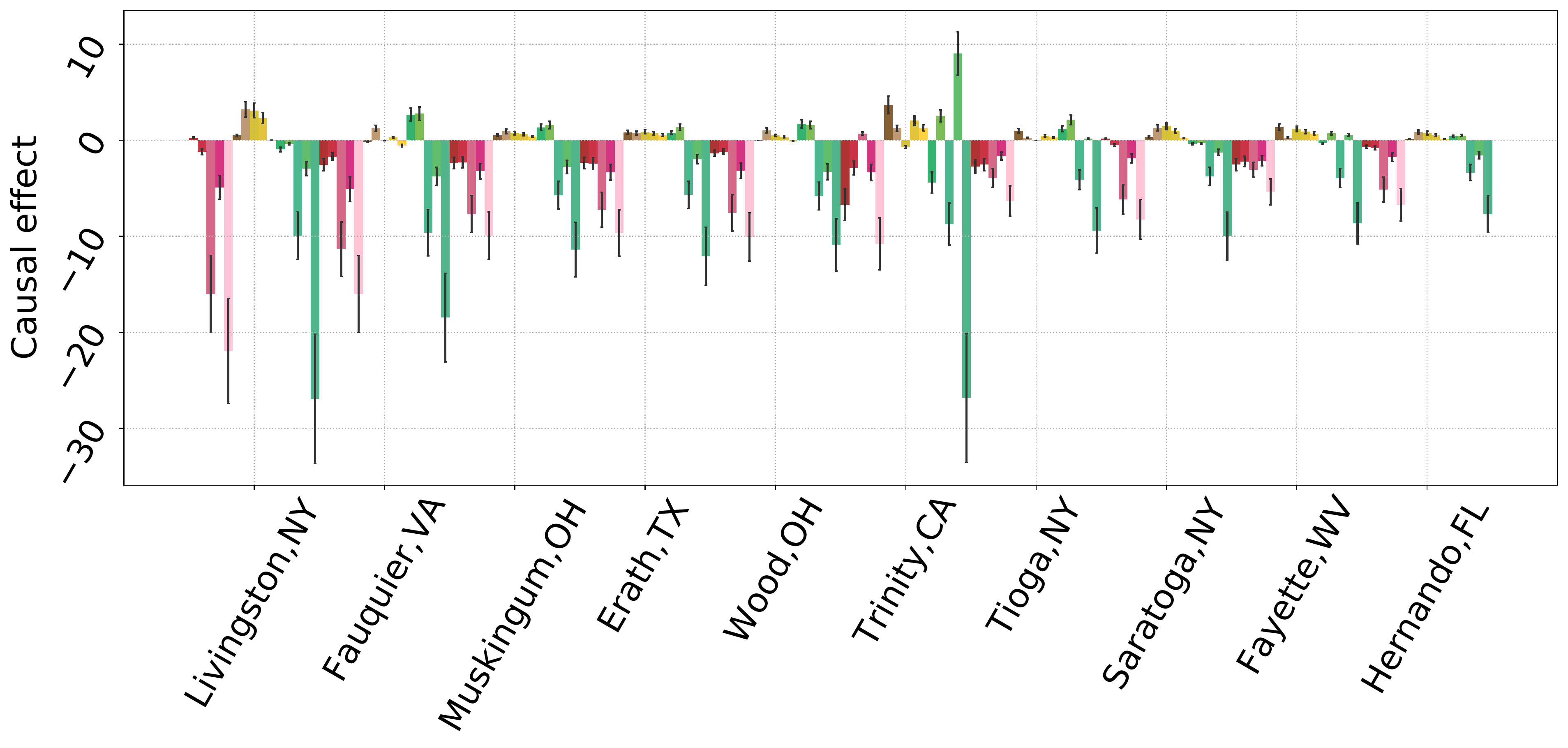}
    } 
    \vspace{-1mm}
    \subfloat[Death cases.]{
    	\label{ite_death}
        \includegraphics[width=0.47\textwidth,height=1.5in]{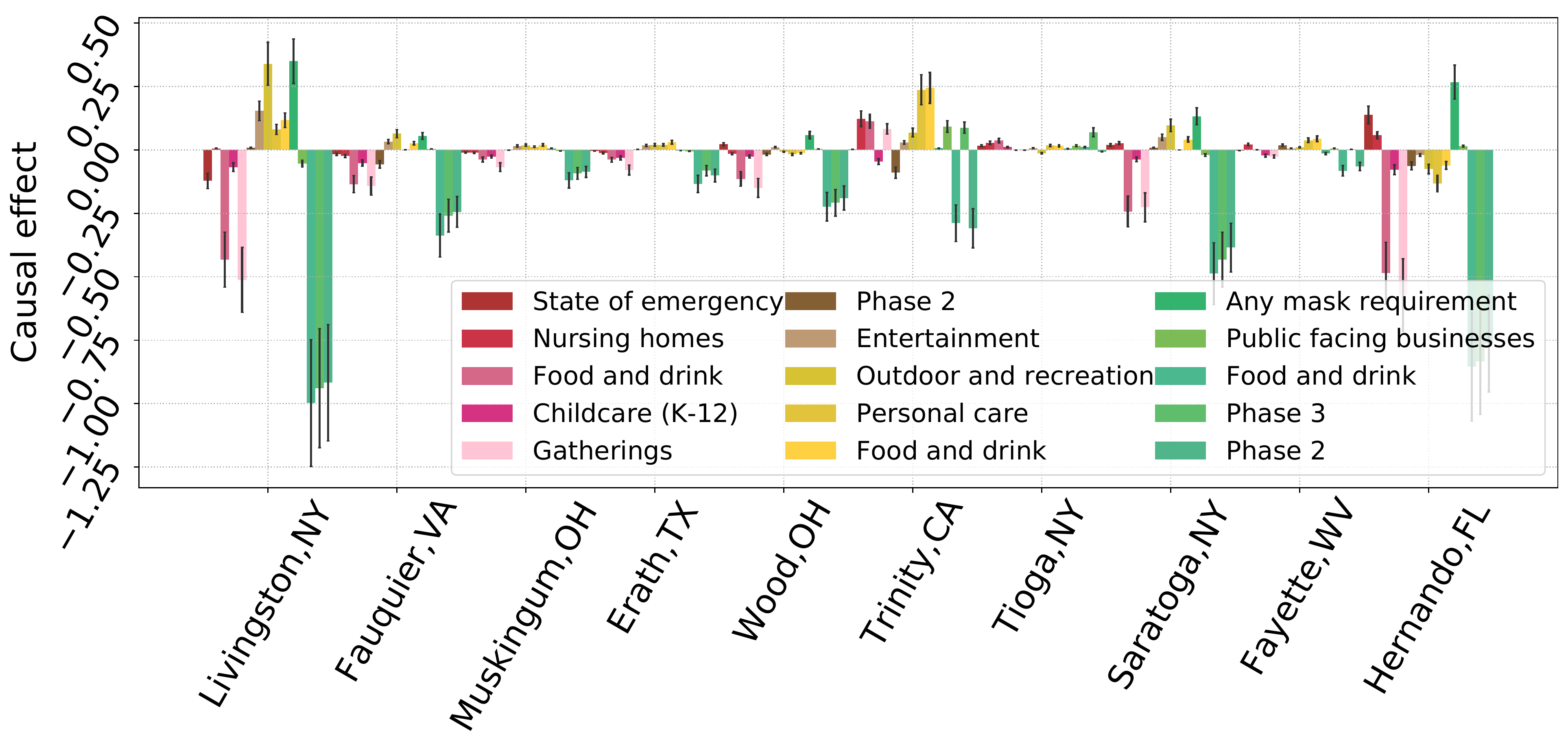}
    }
    \vspace{2mm}
	\caption{Causal effect estimation of different policy types 
	on the outbreak dynamics in different counties. a) Causal effects on the number of confirmed cases; b)  Causal effects on the number of death cases. The red, yellow and green bars correspond to the policy categories of social distancing, reopening, and mask requirement, respectively.}
	\label{fig:ite}
	\vspace{2mm}
\end{figure*}

Furthermore, we zoom into the county-level, and assess the individual treatment effects (ITEs) of different policy types in these three categories on the outbreak dynamics in each county. In Fig~\ref{fig:ite}, we randomly select 
$10$
counties as examples and show the estimated ITE results of different policy types in each of 
them.
To compare 
counties, each result is calculated from the original ITEs (as defined in Eq.~\ref{eq:ite}) averaged over all the time periods, and then normalized by the number of confirmed cases in the corresponding county at the last time period. From Fig.~\ref{fig:ite}, we observe that the ITEs of the three categories of policies in each county have generally similar patterns as their ATEs over all counties, i.e., the ``social distancing'' and ``mask requirement'' policies are still beneficial for controlling the spread of COVID-19, while the ``reopening'' policies have increased the risk. Besides, the policies have a stronger impact in high-risk locations such as counties in California, New York, and Florida.

\subsection{Prediction of Outbreak Dynamics and Presence of Policies with Learned Confounders}
As the counterfactual outcomes are difficult to 
obtain, it is hard to quantitatively evaluate whether our proposed framework renders satisfactory results of causal effect estimation. Fortunately, the prediction performance of outbreak dynamics (outcome) and presence of policies (treatment) based on the learned confounder representations can serve as a good indicator to show the capability of the proposed framework in capturing the  confounders. Through such prediction performance, the effectiveness of causal effect estimation can also be implicitly assessed. Here we compare the prediction performance and causal effect estimation results of our proposed framework with multiple baselines, including the state-of-the-art causal effect estimation methods, as well as some variants of our proposed framework for ablation study. The compared baselines are described as below: (1) \textbf{Naive estimation of ATE} \cite{rubin2005bayesian} --- This method estimates the causal effect of each policy by simply taking the difference between the average values of the observed outcomes in the treated group and the control group. This method may suffer from confounding bias. (2) \textbf{Outcome regression} \cite{rubin2005bayesian} --- Outcome regression is a commonly used method in causal effect estimation. It takes the covariates as input to predict the potential outcomes under each treatment assignment. We implement it with linear regression. (3) \textbf{Difference-in-differences (DID)} --- DID estimates the causal effect by comparing the average change of the outcome during each time period in the treated group with the change in the control group. It is based on the parallel trend assumption \cite{goodman2020using}. (4) \textbf{Causal effect variational autoencoder (CEVAE)} \cite{CEVAE} --- CEVAE is a deep latent-variable model, which learns representations of confounders as Gaussian distributions from original features, observed treatment assignment, and factual outcome. (5) \textbf{Counterfactual Regression (CFR)} \cite{CFR} --- CFR learns representation for the confounders based on the unconfoundedness assumption \cite{rubin2005bayesian}, and predicts the potential outcomes based on the learned representations. 
(6) \textbf{No network (Ours-NG)} --- In this variant of our framework, we remove the GNN module to disable this variant from utilizing the network structure. (7) \textbf{No temporal (Ours-NT)} --- In this variant, we remove the RNN based memory unit to disable this variant from utilizing the temporal information.

\begin{table}[]
\centering
\caption{
Comparison of the averaged performance of different methods for predicting the number of confirmed/death cases (outcome prediction) and the presence of policies in three categories (treatment prediction; ``SD'': social distancing, ``RO'': reopening, ``MA'': mask requirement).
}
\label{tab:pred}
\begin{tabular}{c|cc|ccc}
\hline
           & \multicolumn{2}{|c|}{RMSE of} & \multicolumn{3}{|c}{Accuracy of} \\
           & \multicolumn{2}{|c|}{outcome prediction}&\multicolumn{3}{|c}{treatment prediction}\\ 
\hline
Methods    & Confirmed          & Death         & SD      & RO     & MA          \\
\hline
Regression & $13145.34$      & $395.83$      & -         & -         & -         \\
CEVAE      & $12236.19$      & $378.25$      & -         & -         & -         \\
CFR        & $12577.42$      & $316.46$      & -         & -         & -         \\
Ours-NT    & $11540.94$      & $287.91$      & $86.32\%$ & $87.18\%$ & $84.35\%$ \\
Ours-NG    & $751.82$      & $62.41$      & $88.14\%$ & $89.24\%$ & $86.93\%$ \\
Ours-dist  & $657.44$      & $53.54$      & $90.11\%$ & $90.12\%$ & $88.56\%$ \\
Ours-mob   & $694.48$      & $55.25$      & $89.32\%$ & $90.05\%$ & $87.38\%$ \\
\hline
\end{tabular}
\vspace{-2mm}
\end{table}

\begin{figure}[!t]
	\vspace{-2mm}
	\centering 
    \subfloat[Confirmed cases.]{
	\label{effect_confirm_MA}
        \includegraphics[width=0.24\textwidth,height=1.15in]{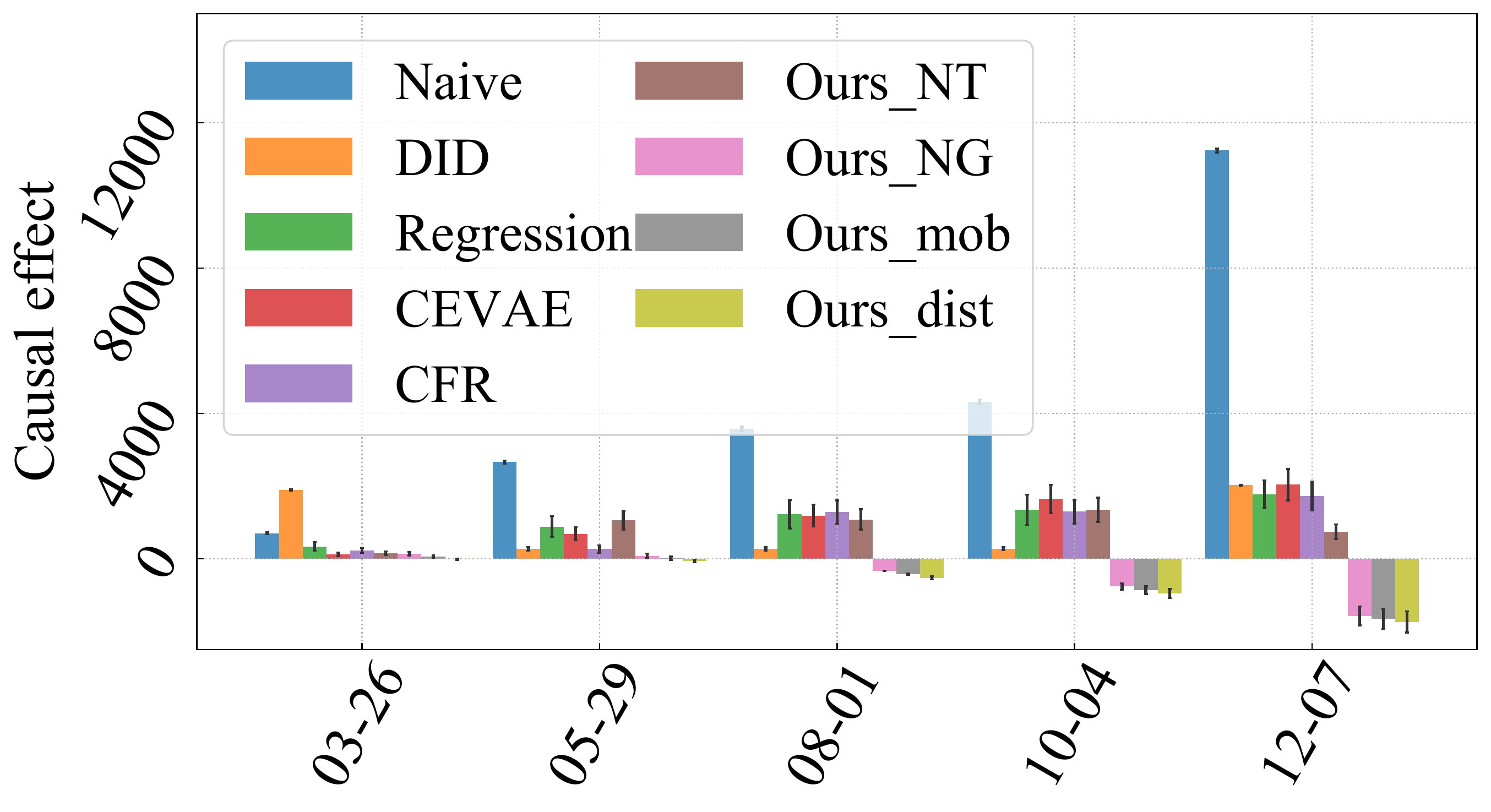}
    } 
    \subfloat[Death cases.]{
    	\label{effect_death_MA}
        \includegraphics[width=0.24\textwidth,height=1.15in]{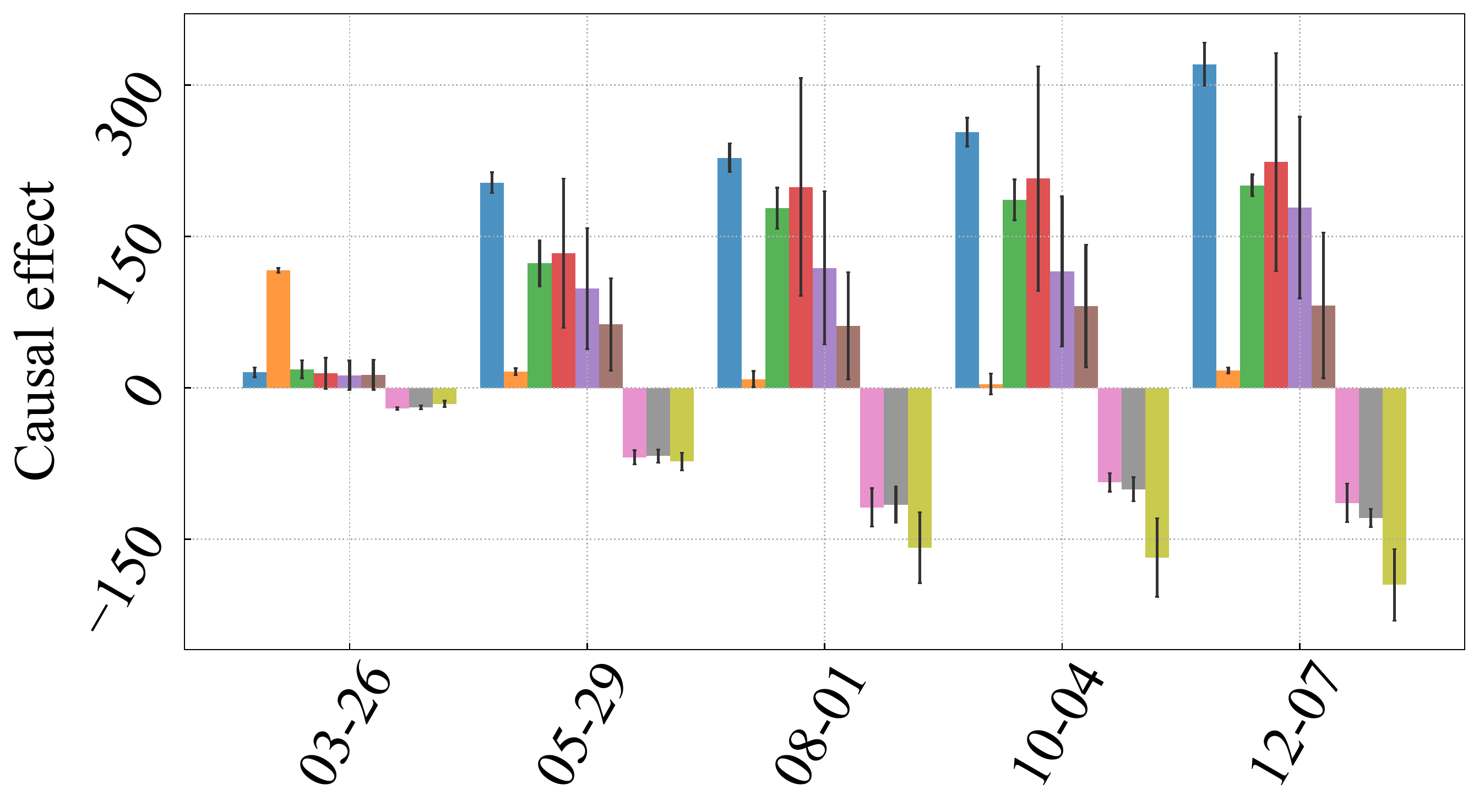}
    }
    \vspace{-2mm}
	\caption{Causal effect estimation of mask-related policy types at different time periods throughout 2020 as computed by different methods. a) Causal effects on confirmed cases; b) Causal effects on death cases. For abbreviations, see text.}  
	\label{fig:confounder}
	\vspace{-5mm}
\end{figure}

To validate the effectiveness of our framework, we first use the prediction results of observed outcomes (including the numbers of confirmed cases and death cases) and treatment assignment (whether the policies are present). We denote the results of our framework based on the distance network and mobility network with postfix -dist and -mob, respectively. As the baseline Outcome regression, CEVAE and CFR cannot handle the temporal information, 
we use them separately at each time period. Table~\ref{tab:pred} shows that our framework can achieve better outcome prediction (i.e., it captures outbreak dynamics better) than the other methods. As these other methods do not model the treatment assignment prediction, we compare our method and its variants in treatment assignment prediction w.r.t. the policy types in different categories. The results show that our framework can achieve high performance in both outcome and treatment assignment prediction, 
indicating
that our framework can 
capture the confounders well, as discussed in \cite{CFR}. By comparing our framework and its two variants, we observe that the network information and temporal information benefits both outcome and treatment assignment prediction, 
suggesting
that this information could help capture more (unobserved) confounders.

\subsection{Causal Effect Estimation Benchmarking}
To further investigate the capability of our framework in capturing the unobserved confounders, we conduct the following case study: we integrate the policies in the mask category as one treatment, and Fig.~\ref{fig:confounder} shows the comparison of the causal effect estimated by different methods. We observe that the results estimated by almost all the baselines stay positive values over time, which conflicts with our common knowledge. 
This phenomenon may result from some unobserved confounders. For example, people in the counties which are under more severe situations (e.g., higher infection rates in itself or neighboring counties) may prone to adopt the mask requirement at an early stage, while other counties may not be so alert to adopt these policies very early. Without effective ways of capturing the confounders (e.g., current situation), a method may incorrectly take the dependency between ``wearing mask'' and ``high number of confirmed cases'' as a causal relationship between them, and lead to such biased estimation.
Here our framework can capture the confounders from the proxies for them including covariates, networks, and historical information, thus the proposed framework can better control for confounding and provide estimation results which are more consistent with common knowledge and existing epidemiological studies of COVID-19~\cite{mitze2020face}. 


%% file: related.tex
\section{Related Work}
The past year has witnessed a surge of works that study the impact of different policies on reducing the spread of COVID-19.
In this section, we focus on 
works in the data science and statistical machine learning areas, 
grouping them 
as follows:
1) non-causal analysis of COVID-19 related policies; 2) causal analysis of COVID-19 related policies; 3) other causal analysis regarding COVID-19.

\noindent \textbf{Non-causal analysis of COVID-19 policies.} Various studies have investigated the COVID-19 related policies \cite{ozili2020covid,capano2020mobilizing}, including their relations with multiple social and economic factors such as household incomes \cite{brewer2020initial}, economics \cite{hevia2020conceptual,schwendicke2020impact}, and vaccination \cite{sharma2020bcg,miller2020correlation}. 
For example, a conceptual framework was proposed in \cite{hevia2020conceptual}, which analyzes how non-pharmaceutical interventions (NPIs) such as social distancing will impact economics. 
Another study \cite{jia2020modeling} conducted correlation analysis and used simulation methods to show the effectiveness of different COVID-19 policies in controlling this pandemic. 
The main contributions of these studies lie in collecting data regarding COVID-19 policies, showing the primary goal, time, and locations of these policies during the pandemic, and using some straightforward statistical methods to analyze these
policies from various aspects. However, these works can only reveal the statistical dependencies between the policies and other factors, rather than identifying the causal effects among them.

\noindent\textbf{Causal analysis of COVID-19 policies.} Beyond 
statistical dependencies, some studies \cite{hsiang2020effect,kristjanpoller2021causal} estimated the causal effect of different policies on COVID-19 dynamics. 
Many works use 
classical causal effect estimation methods based on structural causal models \cite{gencoglu2020causal,feroze2020forecasting,heckerman1999bayesian}, difference-in-differences (DID) \cite{card2000minimum}, or synthetic control \cite{mitze2020face}. 
Among them, those works \cite{feroze2020forecasting} based on Pearl's structural causal models \cite{huang2012pearl} use domain knowledge or causal discovery methods \cite{heckerman1999bayesian} to obtain a complete causal graph
describing
the causal relationships among different variables. However, it is often difficult to obtain a correct and complete causal graph among various factors. 
Of
the methods which do not assume a complete causal graph,
DID methods \cite{koebel2020labor} estimate the causal effect of a treatment (COVID-19 related policies in our case) by comparing the average change of outcome over time in the treated group with the control group. 
Other studies \cite{mitze2020face} that are based on synthetic control methods can account for the effects of evolving confounders over time by constructing a weighted combination of groups used as controls. 
However, despite the contributions of all these works on the insights of COVID-19 policy effect estimation, most of them are either based on the unconfoundedness assumption \cite{rubin2005bayesian} or the parallel trend assumption \cite{abadie2005semiparametric} (i.e., there are no unobserved factors that influence both the cause and the growth trend of the outcome, e.g., the change of the number of confirmed cases 
in
a given month).



\noindent\textbf{Other causal analysis regarding COVID-19.} Besides the causal effect estimation of policies on the COVID-19 outbreak dynamics, there are other causal analyses \cite{carleton2020causal,davies2020evidence,fenton2020covid,sahin2020developing,sodhi2020safety} related to COVID-19 which have yielded interesting insights. A work \cite{gencoglu2020causal} discovered causal relationships between pandemic characteristics (e.g. number of infections and deaths) and public sentiment such as Twitter activity. Other works investigated the causal effect of peoples' behaviors during COVID-19 on different variables, such as how working at home  affects collaboration \cite{yang2020work}, and how peoples' mobility and awareness affect COVID-19 infection \cite{steiger2020causal}. These works have provided interesting insights on analyzing the causal impact of different aspects of COVID-19. Specifically, in this paper, we focus on the causal effect of COVID-19 policies on the numbers of confirmed cases and death cases.